\documentclass{article}

\PassOptionsToPackage{numbers, compress}{natbib}

\usepackage[preprint]{neurips_2025}

\usepackage{xcolor}
\definecolor{cvprblue}{rgb}{0.21,0.49,0.74}
\usepackage[utf8]{inputenc} 
\usepackage[T1]{fontenc}    
\usepackage[pagebackref,breaklinks,colorlinks,citecolor=cvprblue]{hyperref}       
\usepackage{url}            
\usepackage{booktabs}       
\usepackage{amsfonts}       
\usepackage{nicefrac}       
\usepackage{microtype}      
\usepackage{xcolor}         
\usepackage{amssymb}
\usepackage{amsmath}
\usepackage{enumitem}
\usepackage{graphicx}
\usepackage{lineno}
\usepackage{xspace}
\usepackage{algorithm}
\usepackage{algpseudocode}
\usepackage{array}
\usepackage{longtable}
\usepackage{arydshln}
\usepackage{wrapfig}
\usepackage{caption}
\usepackage{multirow}
\usepackage{subfigure}   
\usepackage{listings}

\newcommand{\Ours}{\textsc{Search-R1}\xspace}

\newcommand{\think}[1]{\textcolor{blue}{\texttt{<think>}} #1 \textcolor{blue}{\texttt{</think>}}}
\newcommand{\search}[1]{\textcolor{cyan}{\texttt{<search>}} #1 \textcolor{cyan}{\texttt{</search>}}}
\newcommand{\info}[1]{\textcolor{brown}{\texttt{<information>}} #1 \textcolor{brown}{\texttt{</information>}}}
\newcommand{\answer}[1]{\textcolor{purple}{\texttt{<answer>}} #1 \textcolor{purple}{\texttt{</answer>}}}

\newcommand{\se}{\mathcal{R}}

\title{An Empirical Study on Reinforcement Learning for Reasoning-Search Interleaved LLM Agents}

\author{Bowen Jin$^1$\thanks{Part of the work was done while Bowen was a student researcher at Google.}
, Jinsung Yoon$^2$, Priyanka Kargupta$^1$, Sercan Ö. Arık$^2$, Jiawei Han$^1$ \\
$^1$ University of Illinois at Urbana-Champaign \\
$^2$ Google Cloud AI Research \\
\texttt{bowenj4@illinois.edu} \\
}

\begin{document}

\maketitle

\begin{abstract}
Reinforcement learning (RL) has demonstrated strong potential in training large language models (LLMs) capable of complex reasoning for real-world problem solving. 
More recently, RL has been leveraged to create sophisticated LLM-based search agents that adeptly combine reasoning with search engine use.
While the use of RL for training search agents is promising, the optimal design of such agents remains not fully understood. 
In particular, key factors---such as 
(1) reward formulation, 
(2) the choice and characteristics of the underlying LLM, and 
(3) the role of the search engine in the RL process---require further investigation.
In this work, we conduct comprehensive empirical studies to systematically investigate these and offer actionable insights. 
We highlight several key findings: 
format rewards are effective in improving final performance, whereas intermediate retrieval rewards have limited impact; 
the scale and initialization of the LLM (general-purpose vs. reasoning-specialized) significantly influence RL outcomes; 
and the choice of search engine plays a critical role in shaping RL training dynamics and the robustness of the trained agent during inference.
These establish important guidelines for successfully building and deploying LLM-based search agents in real-world applications.
Code is available at \url{https://github.com/PeterGriffinJin/Search-R1}.
\end{abstract}

\section{Introduction}
Large language models (LLMs) \cite{zhao2023survey} have demonstrated exceptional performance across a range of natural language processing tasks, including question answering \cite{liu2024chatqa}, summarization \cite{zhang2024benchmarking}, and open-ended text generation \cite{gomez2023confederacy}.
Recently, inspired by developments such as DeepSeek-R1 \cite{guo2025deepseek}, reinforcement learning (RL) \cite{kaelbling1996reinforcement,sutton1999reinforcement} has been increasingly applied to LLMs to unlock more advanced reasoning capabilities \cite{wei2022chain}.
LLMs trained via RL have shown strong performance in tasks requiring logical reasoning \cite{xie2025logic} and visual understanding \cite{zhan2025vision}, with promising applications emerging in specialized domains such as finance \cite{liu2025fin} and medicine \cite{lai2025med}.
However, these models often remain limited to self-contained reasoning and lack the ability to interact with external environments or leverage external tools.
To address this, recent work has explored using RL to train LLMs as interactive agents, capable of engaging with external environments and invoking tools, as demonstrated in multi-turn game tasks \cite{wang2025ragen} and user interface control tasks \cite{liu2025infigui,xia2025gui}.

A key agentic application for LLMs is \textit{search}, where models decompose complex problems, perform multi-turn reasoning, and iteratively interact with search engines to retrieve relevant information.
Prior work has explored prompt-based approaches \cite{jin2024long,trivedi2022interleaving} and supervised fine-tuning (SFT) methods \cite{asai2023self,schick2023toolformer} to equip LLMs with search capabilities.
However, these approaches face key limitations: LLMs typically lack strong search proficiency from pretraining alone, and SFT requires costly manual annotation of intermediate reasoning trajectories, making it challenging to scale.
In contrast, recent studies \cite{chen2025research,jin2025search,song2025r1,zheng2025deepresearcher} demonstrate that \textit{RL with outcome-based rewards} offers an effective alternative for training LLMs to perform reasoning and search in an interleaved manner---forming what is referred to as an \textit{LLM-based search agent}.
This training paradigm enhances the model’s ability to interact with search engines while eliminating the need for explicit supervision of intermediate reasoning steps, thus enabling scalable and more flexible agent learning.

While recent RL-based methods have demonstrated the potential to train effective LLM-based search agents, several key questions remain underexplored:
(1) \textit{How does reward design affect search agent training?}
Although prior work \cite{jin2025search} shows that outcome-based rewards alone can activate reasoning and search capabilities, it is unclear whether auxiliary rewards such as format rewards (which signal adherence to the agentic action format \cite{guo2025deepseek}) or intermediate retrieval rewards (which iteratively incentivize outcome-relevant retrievals \cite{lin2025rec}) can further enhance performance. 
(2) \textit{How does the backbone LLM influence RL dynamics?}
As suggested by \cite{gandhi2025cognitive}, the choice of the base model is critical. 
Factors such as model scale (\textit{e.g.}, 3B vs. 32B) and type (\textit{e.g.}, general-purpose vs. reasoning-specialized) can significantly impact the learning dynamics.
(3) \textit{How does the search engine choice affect the learned agent?}
This includes understanding how the quality of different search engines influence RL training dynamics and whether the resulting agent remains robust when the retrieval system is changed at inference time.

In this paper, we conduct comprehensive empirical studies to address the aforementioned research questions. 
Our key findings are summarized as follows:
(1) \textit{Reward Design}. We observe that incorporating a format reward significantly improves performance, particularly when training from a base LLM rather than an instruction-tuned one. 
In contrast, intermediate retrieval rewards do not yield consistent performance improvements, suggesting limited utility.
(2) \textit{Underlying LLM Backbone}. General-purpose LLMs outperform reasoning-specialized LLMs in RL settings, likely due to the latter’s weaker instruction-following capabilities at the early stages of training. 
Furthermore, scaling up the backbone model generally enhances final performance, although with diminishing returns.
(3) \textit{Search Engine Choice}. The quality of the search engine used during training strongly influences RL dynamics. 
Training with a non-informative search engine (\textit{e.g.}, random noise) leads the agent to avoid retrieval altogether, while a weak engine (\textit{e.g.}, BM25 \cite{robertson2009probabilistic}) results in frequent but less efficient search calls. 
In contrast, strong engines (\textit{e.g.}, dense retrievers) yield more stable learning. 
At inference time, the search agent is generally robust to diverse retrieval systems, and stronger search engines consistently lead to better downstream performance.

\section{Related Works}

\subsection{Large Language Models and Reinforcement Learning}

RL \citep{kaelbling1996reinforcement, sutton1999reinforcement} offers a principled framework for sequential decision-making, where an agent optimizes its behavior by interacting with an environment and maximizing cumulative rewards. 
In the context of LLM tuning, RL was popularized by Reinforcement Learning from Human Feedback (RLHF) \citep{jin2025llm, kaufmann2023survey, ouyang2022training}, which first trains a reward model from human preference data \citep{lambert2024rewardbench} and then fine-tunes the policy LLM via Proximal Policy Optimization (PPO).
While PPO enables high-quality alignment, it incurs significant computational overhead due to iterative optimization steps.
Recent efforts to strike a better balance include Group Relative Policy Optimization (GRPO) \citep{shao2024deepseekmath}, which removes the dependency on a learned value function by leveraging group-based baseline estimation, and RLOO \citep{ahmadian2024back}, a simplified variant of REINFORCE \citep{williams1992simple} tailored for LLM training. 
More recently, DAPO \citep{yu2025dapo} extends GRPO by introducing four key innovations tailored for large-scale LLM reinforcement learning: clip-higher reward capping, dynamic sampling for adaptive data efficiency, a token-level objective for finer-grained supervision, and overlong reward shaping to handle extended sequences. 
In parallel, VAPO \citep{yuan2025vapo} builds upon PPO by proposing a value-model-augmented framework, incorporating value pretraining, a decoupled Generalized Advantage Estimator (GAE), and an auxiliary language modeling loss on positive examples to improve credit assignment and stability. 
While these advancements have significantly enhanced the scalability and efficiency of RL-based LLM tuning, their application to LLM-driven search and reasoning tasks remains underexplored, highlighting a critical direction for future exploration.

\subsection{Large Language Models as Search Agents}

LLMs \citep{achiam2023gpt,team2024gemini,zhao2023survey} have demonstrated strong reasoning capabilities \citep{guo2025deepseek} but often struggle with hallucinations and insufficient domain-specific knowledge \citep{li2023large,peng2023study}. 
To address these, recent efforts explore integrating LLMs with search engines to enable dynamic access to external knowledge.
A prominent direction is to treat search engines as interactive tools that LLMs can call during inference \citep{schick2023toolformer}. 
This search-as-a-tool paradigm allows models to iteratively formulate queries, retrieve relevant content, and revise their responses based on external evidence \citep{trivedi2022interleaving}.
Prompt-based methods such as IRCoT \citep{trivedi2022interleaving} and ReAct \citep{yao2023react} enable interleaved reasoning and retrieval, while Toolformer \citep{schick2023toolformer} and self-RAG \citep{asai2023self} uses supervised fine-tuning to learn when and how to call a search engine.
However, these methods often depend on high-quality demonstration data, which is difficult to scale. As an alternative, RL offers a scalable and data efficient solution. 
Inspired by \cite{guo2025deepseek}, recent works \citep{chen2025research,jin2025search,song2025r1,zheng2025deepresearcher} show that LLMs can acquire complex reasoning and search behaviors through RL only using outcome-based rewards. 
Despite this promise, there is still a lack of in-depth empirical study of different design choices in RL for LLM search agents.

\section{Preliminary}

\textbf{Reasoning-Search Interleaved LLM Agent (\textit{i.e.}, LLM-based Search Agent)}~\citep{jin2025search,singh2025agentic}.
We consider an agentic LLM that performs interleaved, \textit{multi-turn} reasoning and search engine interactions.
In each iteration, the LLM-based search agent first engages in \textit{reasoning} to analyze the current context and identify what additional information is needed.
It then formulates a search query to \textit{retrieve} relevant external information, which is incorporated into the \textit{context} for subsequent reasoning.
This iterative process continues until the model determines that sufficient information has been gathered to produce a final \textit{answer}.
The overall interaction follows a multi-turn reasoning–search loop:
reasoning $\rightarrow$ search $\rightarrow$ context $\rightarrow$ reasoning $\rightarrow$ search $\rightarrow$ ... $\rightarrow$ reasoning $\rightarrow$ answering.
To facilitate this process \cite{yao2023react}, the reasoning steps are enclosed within \think{}, search queries are wrapped in \search{}, retrieved information is inserted into \info{}, and the final answer is placed within \answer{}.

\textbf{RL for Training an LLM-based Search Agent.}
In~\citep{jin2025search,zheng2025deepresearcher}, they propose an RL objective to explicitly incorporate a search engine $\se$ during optimization for LLM search agent training. The objective is formalized as:
\begin{gather}\label{eq:rl-retriever}
    \max_{\pi_\theta} \mathbb{E}_{x \sim \mathcal{D},\, y \sim \pi_{\theta}(\cdot \mid x; \se)}
    \left[ r_{\phi}(x, y) \right] 
    - \beta \mathbb{D}_{\text{KL}}\left[ \pi_{\theta}(y \mid x; \se) \,\|\, \pi_{\text{ref}}(y \mid x; \se) \right],
\end{gather}
where $\pi_{\theta}$ denotes the trainable policy, $\pi_{\text{ref}}$ is a fixed reference model, $r_{\phi}$ represents the reward function, and $\mathbb{D}_{\text{KL}}$ denotes the KL divergence. Here, $x$ are sampled from the dataset $\mathcal{D}$, and $y$ denote the output sequence interleaving reasoning steps with search engine retrievals. 

In contrast to prior approaches that generate rollouts exclusively from the model $\pi_{\theta}(\cdot \mid x)$~\citep{ouyang2022training,rafailov2023direct}, ~\citep{jin2025search,zheng2025deepresearcher} augment the generation process by interleaving retrievals via $\pi_{\theta}(\cdot \mid x; \se)$, which can be interpreted as $\pi_{\theta}(\cdot \mid x) \bigotimes \se$, where $\bigotimes$ denotes a retrieval-reasoning composition.

The reward function $r_{\phi}$ serves as the primary optimization signal. ~\citep{jin2025search,zheng2025deepresearcher} employ a rule-based reward system focusing exclusively on \textbf{final outcome rewards}, which evaluate the correctness of the final answer. 
In factual reasoning tasks, correctness is assessed using exact string match (EM) evaluation: $r_{\phi}(x, y) = \text{EM}(a_\text{pred}, a_\text{gold})$,
where $a_\text{pred}$ is the predicted final answer extracted from the model's response $y$, and $a_\text{gold}$ is the ground-truth answer.
In other words,
\begin{gather}
r_{\phi}(x, y) = 
\begin{cases}
    1              & \text{if } a_{\text{pred}} = a_{\text{gold}}, \\
    0              & \text{if } a_{\text{pred}} \neq a_{\text{gold}}, \\
\end{cases}
\end{gather}
Although prior methods have demonstrated strong performance, there remains a notable gap in empirical studies systematically evaluating key design choices---specifically, the effectiveness of different reward formulations, the influence of underlying LLM characteristics, and the impact of search engine selection---on the reinforcement learning process for training search agents.

\section{RL Rewards for LLM-based Search Agents}\label{sec:reward-study}

In ~\citep{jin2025search,zheng2025deepresearcher}, an outcome-driven reward (\textit{i.e.}, string exact match) is employed through the RL process to guide the LLM on learning the reasoning and interleaved search engine calling functionality.
However, in search scenarios, the LLMs need to follow a specific format in order to call the search engine (\textit{i.e.}, format reward) and the relevance of the intermediate search results can also guide the LLM on generating the proper queries (\textit{i.e.}, intermediate retrieval reward). 

\subsection{Format Reward}\label{sec:format-reward}

\textbf{Motivation.} 
When training an LLM-based search agent capable of reasoning and invoking external search engines, it is common to adopt the reasoning-action-observation workflow \cite{yao2023react}, where relevant content is wrapped within special tokens such as \think{}, \search{}, and \info{}.
For instance, if the LLM fails to correctly format its search queries using \search{}, it cannot successfully trigger the search engine and retrieve the external information needed for problem solving.
Thus, adhering to the prescribed format is critical for ensuring the effectiveness of the search agent.
In this section, we explore how incorporating a format reward influences the RL training process of a search agent.

\textbf{Experimental Design.} 
In addition to the outcome reward defined in ~\citep{jin2025search,zheng2025deepresearcher}, we introduce a format reward, resulting in the final reward function $r_{\phi}(x, y)$:
\begin{gather}
r_{\phi}(x, y) = 
\begin{cases}
    1              & \text{if } a_{\text{pred}} = a_{\text{gold}} \land f_{\text{format}}(y) = \text{True}, \\
    1 - \lambda_f    & \text{if } a_{\text{pred}} = a_{\text{gold}} \land f_{\text{format}}(y) = \text{False}, \\
    \lambda_f        & \text{if } a_{\text{pred}} \neq a_{\text{gold}} \land f_{\text{format}}(y) = \text{True}, \\
    0              & \text{if } a_{\text{pred}} \neq a_{\text{gold}} \land f_{\text{format}}(y) = \text{False}, \\
\end{cases}
\end{gather}
where $f_\text{format}(\cdot)$ verifies whether the response $y$ follows the correct reasoning-action-observation format, including the appropriate use of special tokens.
We assign a reward of $\lambda_f$ when the LLM generates an incorrect answer in the correct format, and a reward of $1 - \lambda_f$ when the answer is correct but the format is incorrect.
Details of the $f_\text{format}(\cdot)$ implementation are provided in Appendix~\ref{apx:sec:format}.
We follow \cite{jin2025search} for the training and testing datasets and use exact match as the outcome reward.
Detailed experimental settings can be found in Appendix \ref{apx:setting1}.

\begin{table}[t]
    \centering
    \scriptsize
    \setlength{\tabcolsep}{4pt}
    \renewcommand{\arraystretch}{1.2}
    \caption{Empirical study of the format reward. \textit{Outcome only} refers to the RL variant with only the outcome reward. Base/Instruct refer to the version of the underlying LLM. $\lambda_f=0.2$ for 3B/14B and $\lambda_f=0.4$ for 7B. The best performance is set in bold. $^\dagger/^\star$ represents in-domain/out-domain datasets.}\label{tab:format}
    \begin{tabular}{lccccccccc}
        \toprule
        & \textbf{Methods} & \multicolumn{3}{c}{\textbf{General QA}} & \multicolumn{4}{c}{\textbf{Multi-Hop QA}} \\
        \cmidrule(lr){3-5} \cmidrule(lr){6-9}
         & & \textbf{NQ$^\dagger$} & \textbf{TriviaQA$^\star$} & \textbf{PopQA$^\star$} & \textbf{HotpotQA$^\dagger$} & \textbf{2wiki$^\star$} & \textbf{Musique$^\star$} & \textbf{Bamboogle$^\star$} & \textbf{Avg.} \\
        \midrule
        \multicolumn{9}{l}{\textbf{Qwen2.5-3B-Base/Instruct}} \\
        \hdashline
        \multirow{4}{*}{PPO} 
        & Outcome only (base)  & 0.406 & 0.587 & 0.435 & 0.284 & 0.273 & 0.049 & 0.088 & 0.303  \\
        & \ \ \ \  w. format reward & \textbf{0.428} & \textbf{0.607} & \textbf{0.459} & \textbf{0.371} & \textbf{0.387} & \textbf{0.150} & \textbf{0.323} & \textbf{0.389} \\
        \cdashline{2-10}
        & Outcome only (instruct)  & 0.341 & 0.545 & 0.378 & 0.324 & \textbf{0.319} & 0.103 & 0.264 & 0.325  \\
        & \ \ \ \  w. format reward & \textbf{0.356} & \textbf{0.557} & \textbf{0.393} & \textbf{0.327} & 0.314 & \textbf{0.122} & \textbf{0.266} & \textbf{0.334} \\
        \hdashline
        \multirow{4}{*}{GRPO} 
        & Outcome only (base)  & 0.421 & 0.583 & 0.413 & 0.297 & 0.274 & 0.066 & 0.128 & 0.312  \\
        & \ \ \ \  w. format reward & \textbf{0.429} & \textbf{0.602} & \textbf{0.435} & \textbf{0.372} & \textbf{0.383} & \textbf{0.148} & \textbf{0.307} & \textbf{0.382} \\
        \cdashline{2-10}
        & Outcome only (instruct)  & \textbf{0.397} & \textbf{0.565} & \textbf{0.391} & \textbf{0.331} & \textbf{0.310} & \textbf{0.124} & 0.232 & \textbf{0.336}  \\
        & \ \ \ \  w. format reward & 0.346 & 0.552 & 0.371 & 0.297 & 0.300 & 0.098 & \textbf{0.266} & 0.319 \\
        \midrule
        \multicolumn{9}{l}{\textbf{Qwen2.5-7B-Base/Instruct}} \\
        \hdashline
        \multirow{4}{*}{PPO} 
        & Outcome only (base)  & 0.480 & 0.638 & 0.457 & 0.433 & 0.382 & \textbf{0.196} & \textbf{0.432} & 0.431  \\
        & \ \ \ \  w. format reward & \textbf{0.488} & \textbf{0.644} & \textbf{0.469} & \textbf{0.436} & \textbf{0.412} & 0.187 & 0.403 & \textbf{0.434} \\
        \cdashline{2-10}
        & Outcome only (instruct)  & \textbf{0.393} & \textbf{0.610} & 0.397 & 0.370 & \textbf{0.414} & 0.146 & 0.368 & \textbf{0.385} \\
        & \ \ \ \  w. format reward & 0.383 & 0.593 & \textbf{0.399} & \textbf{0.376} & 0.317 & \textbf{0.151} & \textbf{0.371} & 0.370 \\
        \hdashline
        \multirow{4}{*}{GRPO} 
        & Outcome only (base)  & 0.395 & 0.560 & 0.388 & 0.326 & 0.297 & 0.125 & 0.360 & 0.350 \\
        & \ \ \ \  w. format reward & \textbf{0.458} & \textbf{0.632} & \textbf{0.442} & \textbf{0.412} & \textbf{0.404} & \textbf{0.180} & \textbf{0.411} & \textbf{0.420} \\
        \cdashline{2-10}
        & Outcome only (instruct)  & \textbf{0.429} & \textbf{0.623} & \textbf{0.427} & \textbf{0.386} & \textbf{0.346} & \textbf{0.162} & \textbf{0.400} & \textbf{0.396} \\
        & \ \ \ \  w. format reward & 0.393 & 0.609 & 0.397 & 0.367 & 0.344 & 0.147 & 0.387 & 0.378 \\
        \midrule
        \multicolumn{9}{l}{\textbf{Qwen2.5-14B-Base/Instruct}} \\
        \hdashline
        \multirow{4}{*}{PPO} 
        & Outcome only (base)  &  0.486 & 0.676 & \textbf{0.480} & \textbf{0.468} & \textbf{0.470} & \textbf{0.241} & \textbf{0.528} & \textbf{0.479 }\\
        & \ \ \ \  w. format reward & \textbf{0.499} & \textbf{0.680} & 0.472 & 0.452 & 0.431 & 0.215 & 0.468 & 0.459 \\
        \cdashline{2-10}
        & Outcome only (instruct) & 0.424 & 0.660 & 0.442 & 0.436 & 0.379 & 0.210 & 0.480 & 0.433  \\
        & \ \ \ \  w. format reward & \textbf{0.449} & \textbf{0.682} & \textbf{0.466} & \textbf{0.447} & \textbf{0.422} & \textbf{0.224} & \textbf{0.500} & \textbf{0.456} \\
        \hdashline
        \multirow{4}{*}{GRPO} 
        & Outcome only (base)  & 0.415 & 0.680 & 0.488 & 0.451 & 0.461 & 0.230 & 0.508 & 0.462 \\
        & \ \ \ \  w. format reward & \textbf{0.500} & \textbf{0.693} & \textbf{0.500} & \textbf{0.481} & \textbf{0.488} & \textbf{0.261} & \textbf{0.516} & \textbf{0.491} \\
        \cdashline{2-10}
        & Outcome only (instruct)  & 0.482 & 0.667 & 0.434 & 0.429 & 0.424 & 0.191 & 0.492 & 0.446 \\
        & \ \ \ \  w. format reward & \textbf{0.488} & \textbf{0.677} & \textbf{0.482} & \textbf{0.455} & \textbf{0.470} & \textbf{0.211} & \textbf{0.516} & \textbf{0.471} \\
        \bottomrule
    \end{tabular}
\end{table}

\begin{figure}[t]
    \centering
    \subfigure[Training reward ($\lambda_f$)]{%
        \includegraphics[width=0.24\textwidth]{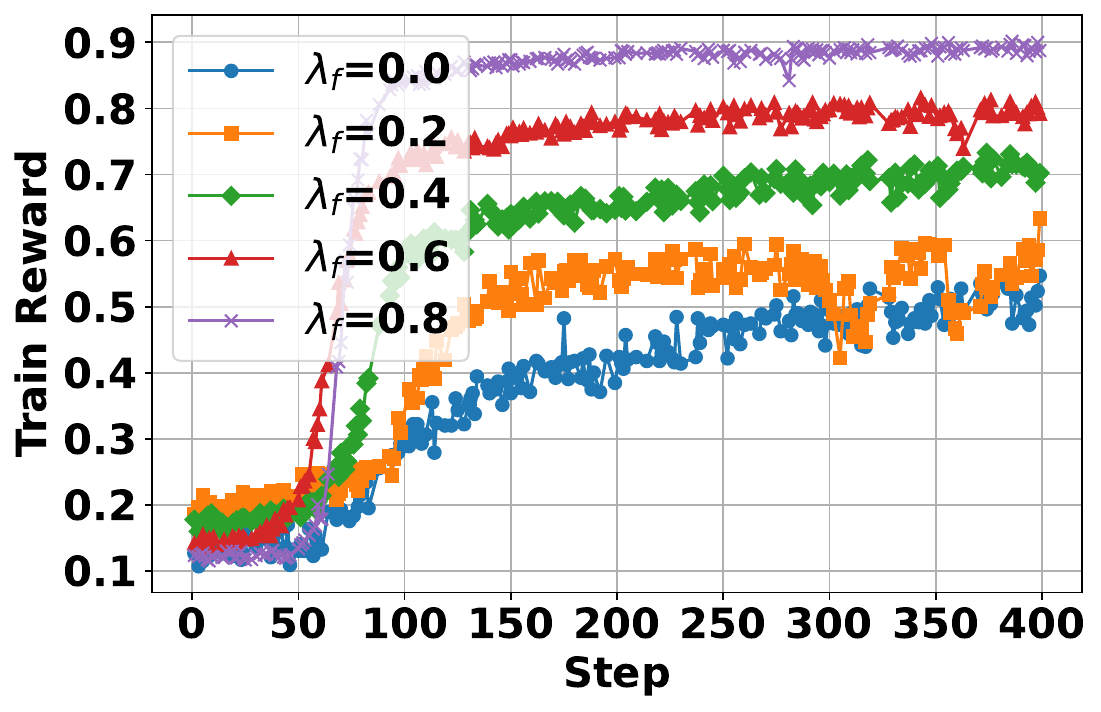}
    }
    \subfigure[$\lambda_f$ sensitivity analysis]{%
        \includegraphics[width=0.24\textwidth]{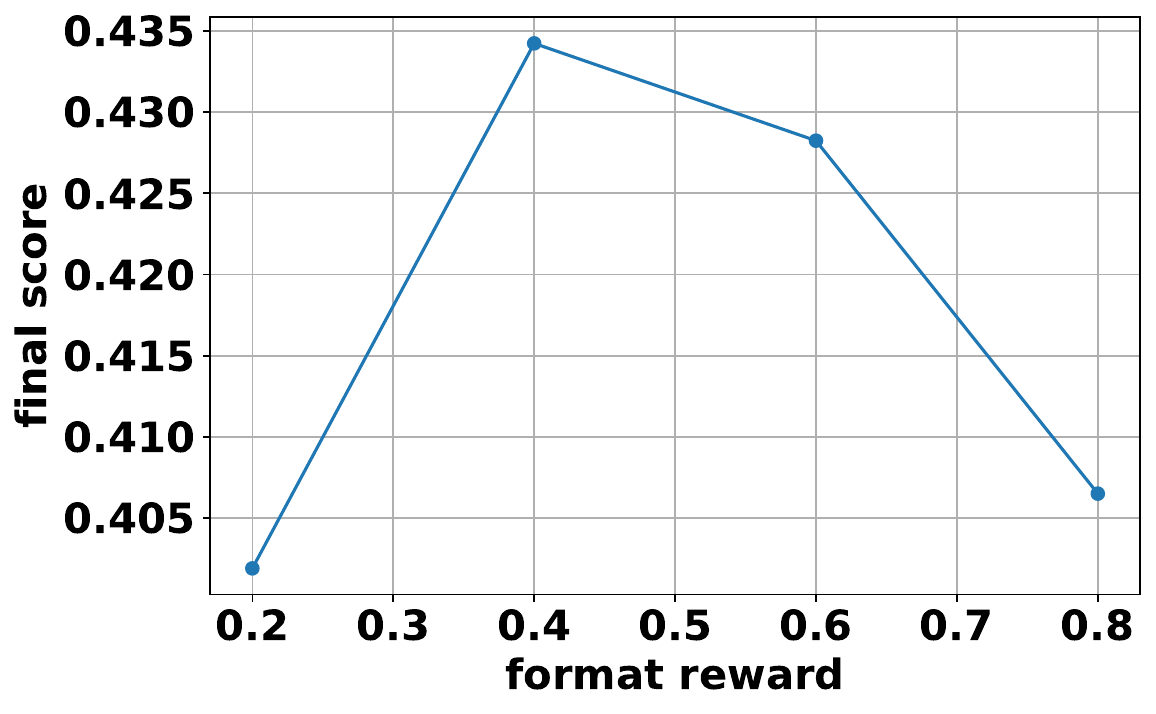}
    }
    \subfigure[Training reward ($\lambda_r$)]{%
        \includegraphics[width=0.23\textwidth]{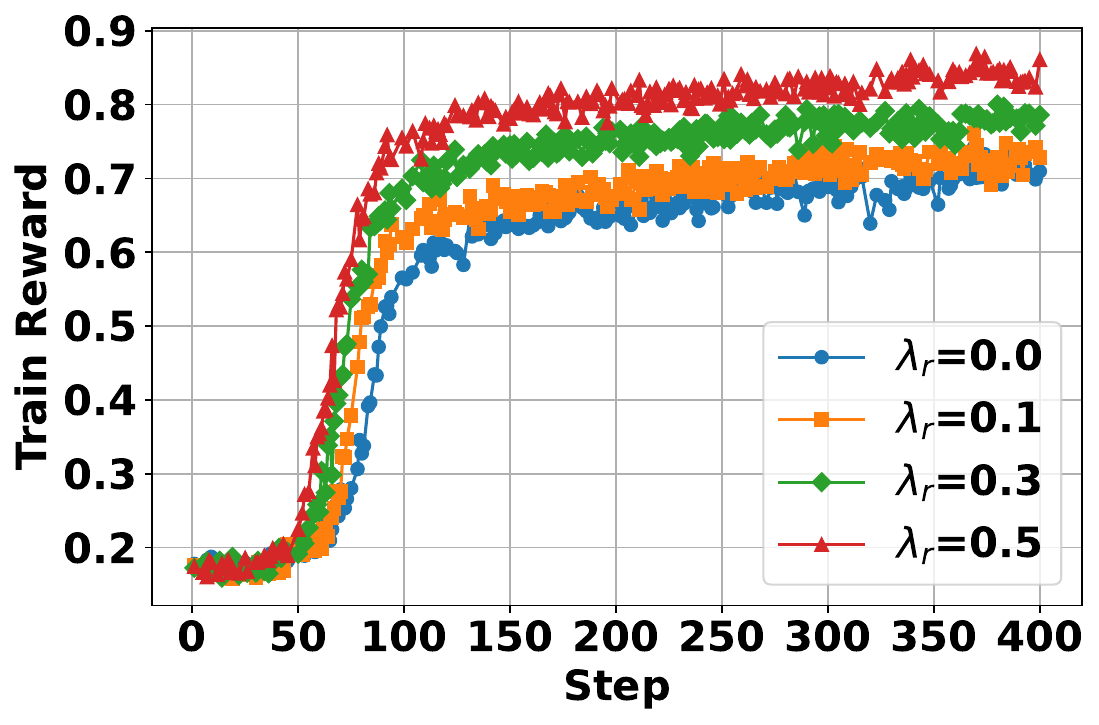}
    }
    \subfigure[$\lambda_r$ sensitivity analysis]{%
        \includegraphics[width=0.23\textwidth]{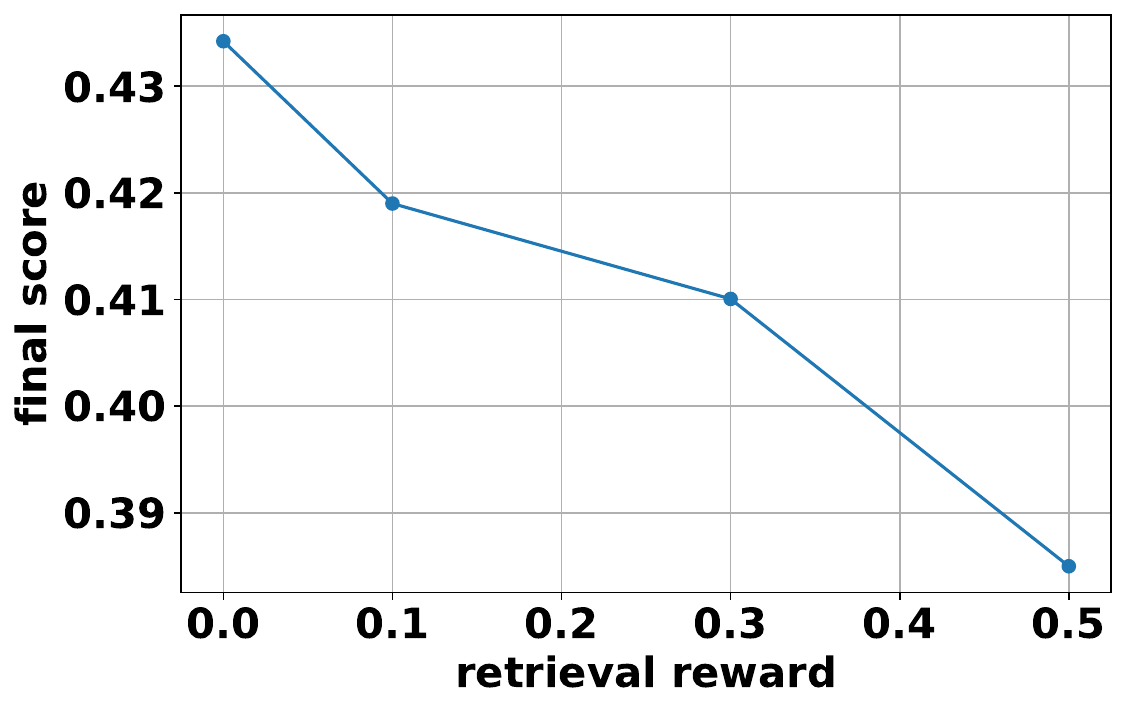}
    }
    \caption{
Empirical analyses on format reward and intermediate retrieval reward.
(a) \textbf{Training reward curves} with varying \textbf{format reward} scaling factors ($\lambda_f$); larger $\lambda_f$ values lead to faster convergence.
(b) \textbf{Impact of $\lambda_f$} on final model performance; a small $\lambda_f$ is ineffective, while an excessively large $\lambda_f$ may cause overfitting to format reward.
(c) \textbf{Training reward} curves under different \textbf{intermediate retrieval reward} scaling factors ($\lambda_r$); varying $\lambda_r$ has limited effect on learning dynamics.
(d) \textbf{Effect of $\lambda_r$} on final model performance; increasing $\lambda_r$ degrades performance, suggesting that intermediate retrieval rewards are unnecessary, as the outcome reward sufficiently encourages effective query formulation.
(LLM: Qwen2.5-7B-Base; RL Algorithm: PPO)
}\label{fig:format-ret-reward}
\end{figure}

\textbf{Results.} 
Table~\ref{tab:format} reports results across various datasets, LLM sizes, and RL algorithms.
Detailed studies on $\lambda_f$ using Qwen2.5-7B-Base and PPO are presented in Figures~\ref{fig:format-ret-reward}(a) and (b).
We summarize the key findings as follows:
(1) Adding a format reward consistently improves final model performance, particularly for base LLMs. 
This is because base LLMs lack strong instruction-following capabilities for search engine invocation, and the format reward helps mitigate this limitation.
(2) Format reward accelerates RL convergence; larger $\lambda_f$ values lead to faster convergence by explicitly guiding the model to issue correctly formatted search queries and interpret results effectively.
(3) The choice of $\lambda_f$ significantly impacts final performance. While a small $\lambda_f$ is ineffective, an excessively large $\lambda_f$ may cause overfitting, ultimately degrading final performance.

\subsection{Intermediate Retrieval Reward}\label{sec:retrieval-reward}

\textbf{Motivation.} 
Beyond the outcome reward, which directly evaluates the correctness of the final answer after multiple search interactions, it is possible to incorporate intermediate retrieval rewards that assess the quality of the retrieved documents during each search step \cite{lin2025rec}.
By assigning positive rewards to cases where relevant information is retrieved, the LLM can be encouraged to generate higher-quality queries that yield more relevant retrieval results \cite{lin2025rec}.
We investigate whether introducing intermediate retrieval rewards benefits the RL training process of LLM-based search agents.

\textbf{Experimental Design.} 
Building upon the outcome reward from~\citep{jin2025search,zheng2025deepresearcher} and the format reward introduced in Section~\ref{sec:format-reward}, we incorporate a retrieval correctness component, resulting in the following final reward function $r_{\phi}(x, y)$:
\begin{gather}
r_{\phi}(x, y) = 
\begin{cases}
    1              & \text{if } a_{\text{pred}} = a_{\text{gold}} \land f_{\text{format}}(y) = \text{True}, \\
    1 - \lambda_f    & \text{if } a_{\text{pred}} = a_{\text{gold}} \land f_{\text{format}}(y) = \text{False}, \\
    \lambda_f + \lambda_r       & \text{if } a_{\text{pred}} \neq a_{\text{gold}} \land f_{\text{format}}(y) = \text{True} \land f_{\text{ret}}(y) = \text{True}, \\
    \lambda_f        & \text{if } a_{\text{pred}} \neq a_{\text{gold}} \land f_{\text{format}}(y) = \text{True} \land f_{\text{ret}}(y) = \text{False}, \\
    0              & \text{if } a_{\text{pred}} \neq a_{\text{gold}} \land f_{\text{format}}(y) = \text{False}, \\
\end{cases}
\end{gather}
where $f_\text{ret}(\cdot)$ determines whether the retrieved documents are relevant to the ground truth answer.
The retrieved information can only be extracted when the rollout sequence follows the desired format, as described in Section~\ref{sec:format-reward} and the purpose of the intermediate retrieval reward is to provide a positive learning signal even when the final answer is incorrect. Thus, we introduce an additional reward term $\lambda_r$ when $a_{\text{pred}} \neq a_{\text{gold}} \land f_{\text{format}}(y) = \text{True}$.
In our experiments, we focus on short-form QA datasets, so we apply substring exact match as $f_\text{ret}(\cdot)$, following \cite{jin2024long,lin2023ra}, 
to evaluate whether ground truth appears in retrieved passages.
Under this setting, even if LLM fails to generate correct final answer, it can still receive a positive reward for issuing effective queries that retrieve relevant documents.
Detailed experimental settings can be found in Appendix \ref{apx:setting2}.

\begin{table}[t]
    \centering
    \scriptsize
    \setlength{\tabcolsep}{4pt}
    \renewcommand{\arraystretch}{1.2}
    \caption{Study of the intermediate retrieval reward. $\lambda_r=0.1$. The best performance is set in bold. $^\dagger/^\star$ represents in-domain/out-domain datasets.}\label{tab:ret-reward}
    \begin{tabular}{llcccccccc}
        \toprule
        & \textbf{Methods} & \multicolumn{3}{c}{\textbf{General QA}} & \multicolumn{4}{c}{\textbf{Multi-Hop QA}} \\
        \cmidrule(lr){3-5} \cmidrule(lr){6-9}
         & & \textbf{NQ$^\dagger$} & \textbf{TriviaQA$^\star$} & \textbf{PopQA$^\star$} & \textbf{HotpotQA$^\dagger$} & \textbf{2wiki$^\star$} & \textbf{Musique$^\star$} & \textbf{Bamboogle$^\star$} & \textbf{Avg.} \\
        \midrule
        \multicolumn{9}{l}{\textbf{Qwen2.5-3B-Base}} \\
        \hdashline
        \multirow{2}{*}{PPO} 
        & w.o. retrieval reward & \textbf{0.428} & \textbf{0.607} & \textbf{0.459} & \textbf{0.371} & \textbf{0.387} & \textbf{0.150} & \textbf{0.323} & \textbf{0.389}  \\
        & w. retrieval reward & 0.405 & 0.567 & 0.407 & 0.326 & 0.330 & 0.104 & 0.242 & 0.340 \\
        \hdashline
        \multirow{2}{*}{GRPO} 
        & w.o. retrieval reward & 0.429 & 0.602 & \textbf{0.435} & 0.372 & \textbf{0.383} & \textbf{0.148} & 0.307 & 0.382  \\
        & w. retrieval reward & \textbf{0.434} & \textbf{0.605} & 0.433 & \textbf{0.379} & 0.378 & 0.142 & \textbf{0.323} & \textbf{0.385} \\
        \midrule
        \multicolumn{9}{l}{\textbf{Qwen2.5-7B-Base}} \\
        \hdashline
        \multirow{2}{*}{PPO} 
        & w.o. retrieval reward & \textbf{0.488} & \textbf{0.644} & \textbf{0.469} & 0.436 & \textbf{0.412} & \textbf{0.187} & \textbf{0.403} & \textbf{0.434} \\
        & w. retrieval reward & 0.472 & 0.629 & 0.452 & 0.436 & 0.402 & 0.180 & 0.363 & 0.419 \\
        \hdashline
        \multirow{2}{*}{GRPO} 
        & w.o. retrieval reward &  \textbf{0.458} & \textbf{0.632} & 0.442 & 	0.412 & \textbf{0.404} & \textbf{0.180} & \textbf{0.411} & \textbf{0.420}  \\
        & w. retrieval reward & 0.453 & 0.628 & \textbf{0.450} & \textbf{0.416} & 0.375 & 0.164 & 0.387 & 0.410 \\
        \bottomrule
    \end{tabular}
\end{table}

\textbf{Results.} 
Performance comparisons with and without intermediate retrieval rewards are presented in Table~\ref{tab:ret-reward}.
The effect of varying $\lambda_r$ is illustrated in Figures~\ref{fig:format-ret-reward}(c) and (d).
Key observations include:
(1) Adding intermediate retrieval rewards does not significantly improve final performance for either PPO or GRPO.
This may be attributed to the outcome reward already providing sufficient learning signal for generating effective queries, as a successful search engine call that retrieves relevant information directly contributes to producing the correct answer and receiving a positive reward.
In contrast, the substring EM-based intermediate retrieval reward may overly constrains the retrieval trajectory and thus deviates the naturally learned trajectory from the outcome reward.
(2) Varying $\lambda_r$ has limited impact on learning dynamics. Increasing $\lambda_r$ consistently leads to degraded performance, suggesting that intermediate retrieval rewards are unnecessary.
The outcome reward alone is sufficient to encourage effective query formulation and downstream task success.

\section{The Impact of Underlying Backbone LLM}\label{sec:backbone-llm}
In this section, we study how the choice of the LLM influences RL training for LLM-based search agents. 
Our investigation centers on two key characteristics of the base LLM:
(1) type (\textit{i.e.}, general-purpose vs. reasoning-optimized), and
(2) scale (\textit{i.e.}, 3B, 7B, 14B, and 32B).

\subsection{Study of LLM types}\label{sec:llm-type}

\textbf{Motivation.}
Effective training of LLM-based search agents via RL requires the LLM to possess two fundamental capabilities: \textit{instruction following} and \textit{reasoning}. 
Instruction following enables the model to learn how to properly issue search engine calls in the correct format, while reasoning equips the model to analyze retrieved information and solve complex problems. 
However, it remains underexplored whether general-purpose or reasoning-specialized LLMs provide a more suitable foundation for RL-based training.

\begin{table}[t]
    \centering
    \scriptsize
    \setlength{\tabcolsep}{4pt}
    \renewcommand{\arraystretch}{1.2}
    \caption{Performance of general LLM and reasoning LLM trained with RL on search agent task. The best performance is set in bold. $^\dagger/^\star$ represents in-domain/out-domain datasets.}\label{tab:llm-type}
    \begin{tabular}{lcccccccc}
        \toprule
        \textbf{Methods} & \multicolumn{3}{c}{\textbf{General QA}} & \multicolumn{4}{c}{\textbf{Multi-Hop QA}} \\
        \cmidrule(lr){2-4} \cmidrule(lr){5-8}
         & \textbf{NQ$^\dagger$} & \textbf{TriviaQA$^\star$} & \textbf{PopQA$^\star$} & \textbf{HotpotQA$^\dagger$} & \textbf{2wiki$^\star$} & \textbf{Musique$^\star$} & \textbf{Bamboogle$^\star$} & \textbf{Avg.} \\
        \midrule
        \multicolumn{9}{l}{\textbf{DeepSeek-R1-Distill-Qwen-7B}} \\
        \hdashline
        PPO & 0.389 & 0.542 & 0.402 & 0.334 & 0.326 & 0.122 & 0.290 & 0.344  \\
        GRPO &  0.061 & 0.155 & 0.068 & 0.098 & 0.194 & 0.010 & 0.113 & 0.100 \\
        \midrule
        \multicolumn{9}{l}{\textbf{Qwen2.5-7B-Base}} \\
        \hdashline
        PPO & \textbf{0.488} & \textbf{0.644} & \textbf{0.469} & \textbf{0.436} & \textbf{0.412} & \textbf{0.187} & 0.403 & \textbf{0.434} \\
        GRPO &  0.458 & 0.632 & 0.442 & 	0.412 & 0.404 & 0.180 & \textbf{0.411} & 0.420  \\
        \bottomrule
    \end{tabular}
\end{table}

\begin{figure}[t]
    \centering
    \subfigure[Training Reward]{%
        \includegraphics[width=0.23\textwidth]{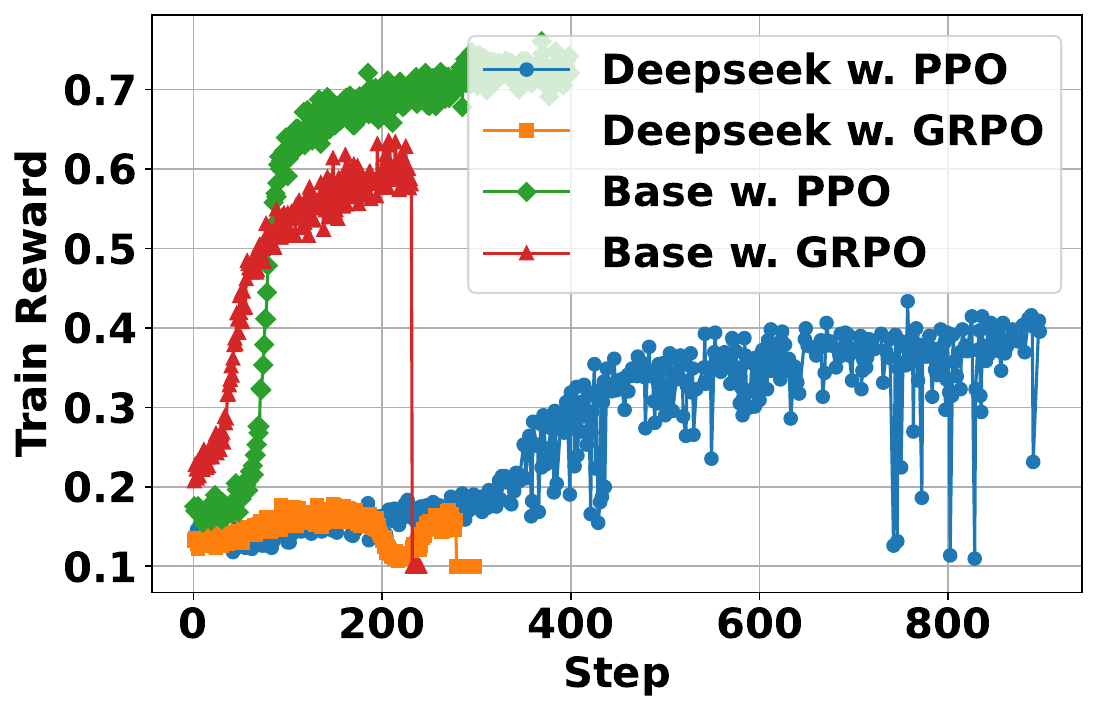}
    }
    \subfigure[\# of Search Calls]{%
        \includegraphics[width=0.23\textwidth]{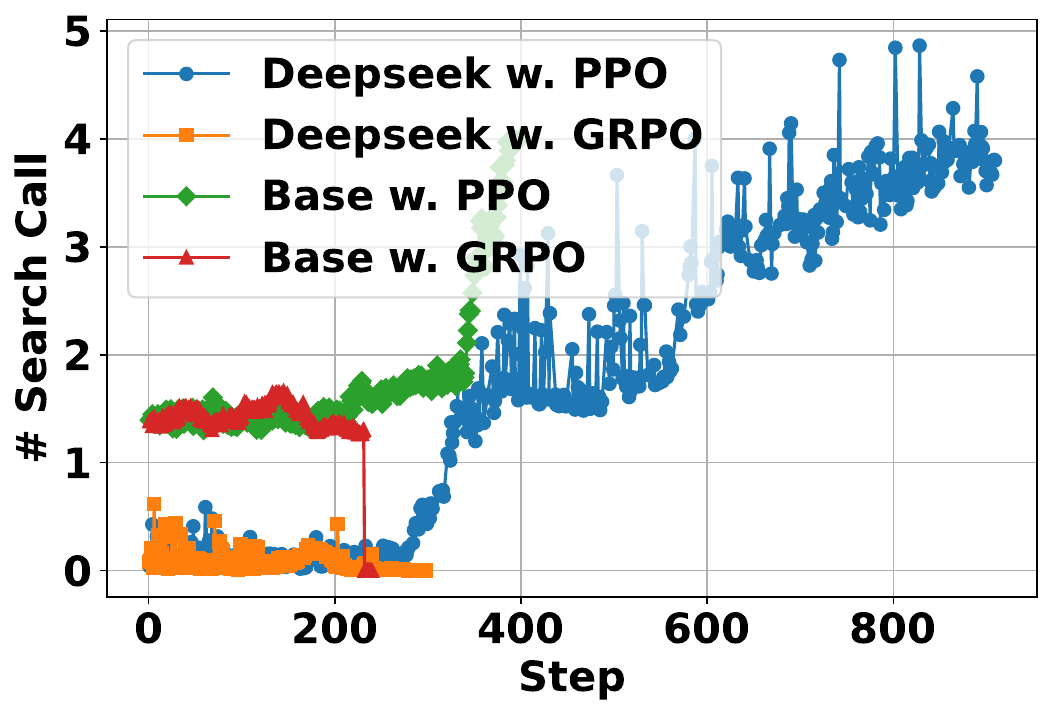}
    }
    \subfigure[Training Reward]{%
        \includegraphics[width=0.24\textwidth]{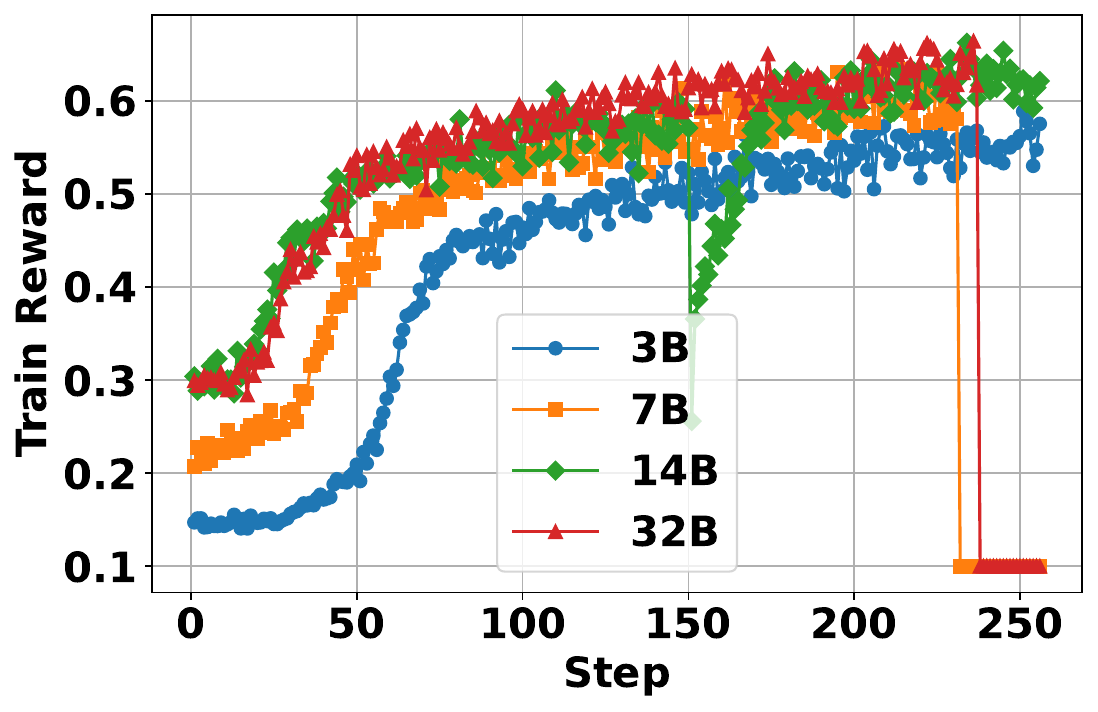}
    }
    \subfigure[Test Accuracy]{%
        \includegraphics[width=0.24\textwidth]{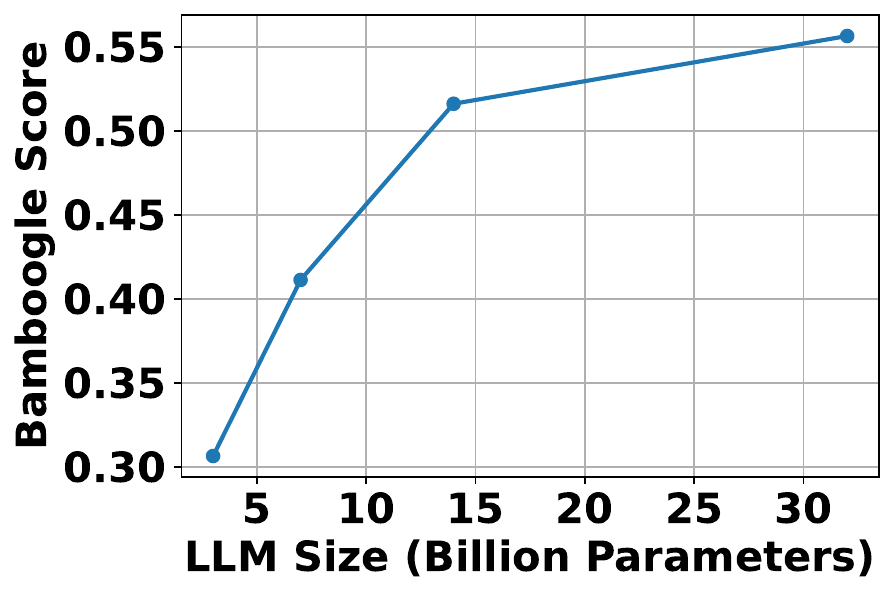}
    }
    \caption{
The study of underlying pretrained LLM for development of LLM-based search agents with RL. 
(a) \textbf{Training reward with different type of LLMs} - general-purpose LLM (Qwen2.5-7B-Base) and reasoning LLM (DeepSeek-R1-Distill-Qwen-7B). We observe that general-purpose LLM performs better than reasoning LLMs with both PPO and GRPO.
(b) \textbf{\# of Search engine calls with different type of LLMs}: General LLM learns to call the search engine faster than the reasoning LLM. This potentially stems from the fact the general LLMs are better for following instructions.
(c) \textbf{Training reward with different size of LLMs}: Larger LLMs can lead to higher training reward.
(d) \textbf{Test accuracy with different size of LLMs}: On the challenging Bamboogle dataset \cite{press2022measuring}, the performance increases consistently as the LLM size increases.
}\label{fig:scaling-LLM}
\end{figure}

\textbf{Experimental Design.}
We follow the experimental setup in \cite{jin2025search} and conduct RL training on two LLM variants:
(1) Qwen2.5-7B-Base \cite{yang2024qwen2}, a general-purpose 7B parameter pretrained LLM, and
(2) DeepSeek-R1-Distill-Qwen-7B \cite{guo2025deepseek}, a 7B reasoning-specialized model distilled from DeepSeek-R1.
Both models are trained under identical conditions to ensure a fair comparison.
Detailed experimental settings and results on 14B LLMs can be found in Appendix \ref{apx:setting3} and \ref{apx:sec:llm-type}.

\textbf{Results.}
The training reward and search engine call frequency curves are presented in Figures~\ref{fig:scaling-LLM}(a) and \ref{fig:scaling-LLM}(b), respectively.
Final performance results are summarized in Table~\ref{tab:llm-type}. We observe the following key findings:
(1) The RL training process is more stable and effective when initialized with the general-purpose LLMs compared to the reasoning-specialized ones. 
This suggests that the general-purpose ones already possess sufficient basic reasoning capabilities to support the search agent task without requiring specialized pretraining.
(2) The reasoning LLM struggles to initiate search engine calls during the early stages of training, leading to insufficient exploration. 
In the absence of positive reward signals from successful rollouts involving search, the model fails to consistently learn to engage with the search engine.
This behavior primarily stems from the reasoning LLMs' limited instruction-following capabilities, which hinder their ability to learn the correct format for invoking the search API.
(3) While the reasoning LLMs eventually learn to perform interleaved reasoning and retrieval when trained with PPO, this progress is slow and gradual. 
In contrast, training with GRPO leads to training collapse. 
We attribute this to PPO’s lower variance and more stable policy updates, which better support the complex exploration required for search-augmented reasoning tasks.

\subsection{The Scale Up of LLM-based Search Agent}\label{sec:llm-scale-up}

\textbf{Motivation.}
Prior work has demonstrated that LLM capabilities improve predictably with increased model size, as described by scaling laws \cite{hoffmann2022training, kaplan2020scaling}. However, it remains unclear whether similar scaling behavior holds when LLMs are further RL-tuned as search agents. 
Specifically, does increasing model size consistently improve the agent's ability to reason and interact with search engines?

\textbf{Experimental Design.}
We evaluate scaling laws using RL with both outcome-based rewards and additional format rewards, as introduced in Section~\ref{sec:format-reward}. 
Following the experimental setup in \cite{jin2025search}, we train Qwen2.5 models of varying sizes (3B, 7B, 14B, 32B) on the NQ and HotpotQA training datasets using the GRPO algorithm. 
We use a fixed learning rate of $5 \times 10^{-7}$, and evaluate on the out-of-distribution Bamboogle dataset.
Detailed experimental settings can be found in Appendix \ref{apx:setting4}.

\textbf{Results.}
As shown in Figure~\ref{fig:scaling-LLM}(c), the training reward consistently improves with increasing LLM size, indicating that larger models are better able to learn effective reasoning and search engine usage. 
The corresponding inference performance is presented in Figure~\ref{fig:scaling-LLM}(d). While test performance also improves with model size, the rate of improvement diminishes. This suggests that the search agent task, unlike pure language modeling, relies less on parametric knowledge stored in large LLMs, and more on effective external information acquisition through retrieval.

\section{Improved LLM-based Search Agents with Stronger Search Engines}\label{sec:search-engine}

The choice of search engine plays a critical role in determining retrieval quality, which in turn influences both the RL training dynamics and the inference-time performance of the LLM-based search agent. 
During training, higher-quality search engines that provide more relevant information can encourage the agent to achieve its objectives with fewer search calls, as the retrieved content more effectively supports reasoning and decision-making. 
In contrast, lower-quality search engines that return less relevant information may lead the agent to either over-rely on its internal knowledge or issue multiple search queries to compensate for inadequate results. 
At inference time, the quality of the retrieved information directly impacts the agent’s ability to generate accurate and useful responses. 
In the following sections, we systematically investigate the effects of search engine choice on both the training and inference stages of search-augmented LLMs.

\begin{table}[t]
    \centering
    \scriptsize
    \setlength{\tabcolsep}{2.6pt}
    \renewcommand{\arraystretch}{1.2}
    \caption{Final performance with different search engine for both training and inference. The best performance is set in bold. (LLM: Qwen2.5-7B-Base; RL: PPO)}\label{tab:train-retriever}
    \begin{tabular}{lcccccccccccccccc}
        \toprule
        \textbf{Engine} & \multicolumn{2}{c}{\textbf{NQ}} & \multicolumn{2}{c}{\textbf{TriviaQA}} & \multicolumn{2}{c}{\textbf{PopQA}} & \multicolumn{2}{c}{\textbf{HotpotQA}} & \multicolumn{2}{c}{\textbf{2wiki}} & \multicolumn{2}{c}{\textbf{Musique}} & \multicolumn{2}{c}{\textbf{Bamboogle}} & \multicolumn{2}{c}{\textbf{Avg.}}  \\
        \cmidrule(lr){2-3} \cmidrule(lr){4-5} \cmidrule(lr){6-7} \cmidrule(lr){8-9} \cmidrule(lr){10-11} \cmidrule(lr){12-13} \cmidrule(lr){14-15} \cmidrule(lr){16-17}
         & \textbf{Recall} & \textbf{EM} & \textbf{Recall} & \textbf{EM} & \textbf{Recall} & \textbf{EM} & \textbf{Recall} & \textbf{EM} & \textbf{Recall} & \textbf{EM} & \textbf{Recall} & \textbf{EM} & \textbf{Recall} & \textbf{EM} & \textbf{Recall} & \textbf{EM}  \\
        \midrule
        Random & 0.000 & 0.237 & 0.000 & 0.494 & 0.000 & 0.177 & 0.000 & 0.217 & 0.000 & 0.269 & 0.000 & 0.058 & 0.000 & 0.234	 & 0.000 & 0.241 \\
        BM25 & 0.216 & 0.341 & 0.445 & 0.607 & 0.255 & 0.322 & 0.273 & 0.404 & \textbf{0.216} & 0.370 & 0.076 & 0.137 & 0.061 & 0.280 & 0.176 & 0.352  \\
        E5 (HNSW) &  0.436 & 0.468 & 0.509 & 0.621 & 0.304 & 0.366 & 0.237 & 0.372 & 0.146 & 0.287 & 0.092 & 0.137 & 0.104 & 0.400	 & 0.261 & 0.379 \\
        E5 (Exact) &  \textbf{0.462} & \textbf{0.481} & \textbf{0.561} & \textbf{0.638} & \textbf{0.423} & \textbf{0.457} & \textbf{0.276} & \textbf{0.433} & 0.198 & \textbf{0.382} & \textbf{0.098} & \textbf{0.196} & \textbf{0.107} & \textbf{0.424} & \textbf{0.304} & \textbf{0.430} \\
        \bottomrule
    \end{tabular}
\end{table}

\begin{figure}[t]
    \centering
    \subfigure[Search engine performance]{%
        \includegraphics[width=0.3\textwidth]{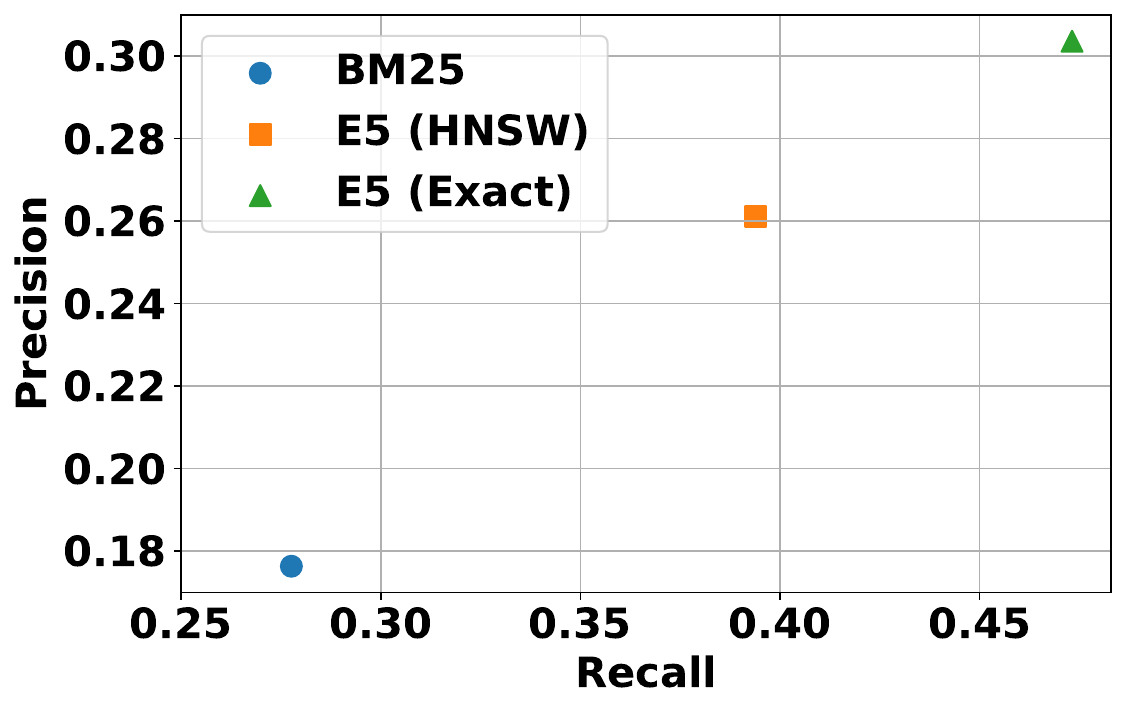}
    }
    \subfigure[Training reward]{%
        \includegraphics[width=0.3\textwidth]{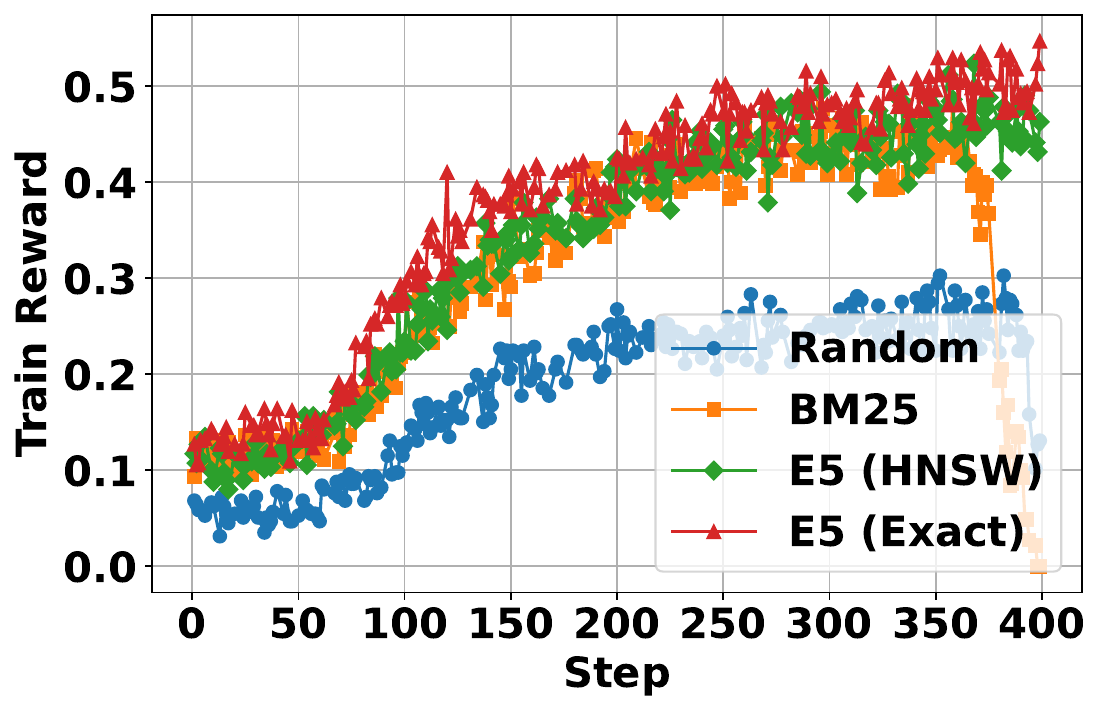}
    }
    \subfigure[\# Search calls]{%
        \includegraphics[width=0.3\textwidth]{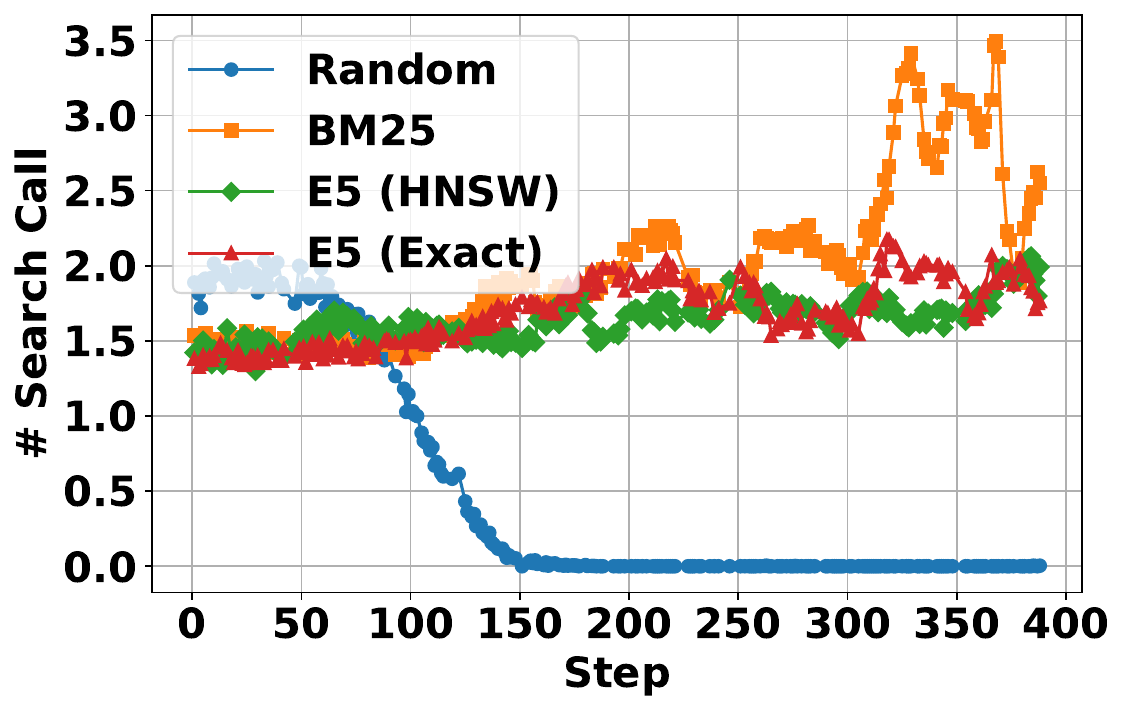}
    }
    \caption{
Effect of Search Engine Choice on RL Training Dynamics.
(a) \textbf{Retrieval Quality Ranking}: E5 (Exact) > E5 (HNSW) > BM25 > Random.
(b) \textbf{Training Stability and Performance}: Stronger search engines (\textit{e.g.}, E5 Exact, E5 + HNSW) lead to more stable training and higher final performance, while weaker engines (\textit{e.g.}, Random, BM25) achieve suboptimal outcomes.
(c) \textbf{Search Engine Usage Behavior}: With Random Noise, the agent quickly learns to avoid using the search engine. With BM25, the agent gradually increases search calls to compensate for limited retrieval quality. With E5, the agent issues search calls more strategically, reflecting more efficient search behavior.
}\label{fig:retriever-training-study}
\end{figure}

\subsection{Training with Different Search Engines.}\label{sec:train-engine}

\textbf{Motivation.}
During RL training, the LLM-based agent learns to interact with the search engine and receive positive reward feedback while it solves problems using the retrieved relevant information. 
A \textit{strong} search engine provides more relevant results, leading to consistent positive outcome rewards. Consequently, the LLM learns to solve problems with fewer search calls. 
In contrast, a \textit{weak} search engine discourages reliance on retrieval or forces the agent to issue multiple search queries to compensate for low-quality results. 
We empirically investigate how different search engines influence the RL training dynamics of an LLM-based search agent.

\textbf{Experimental Design.}
We conduct experiments using the Qwen2.5-7B-Base model as the LLM and Proximal Policy Optimization (PPO) as the RL algorithm. 
Four search engine configurations are explored with the Wikipedia-18 corpus \cite{karpukhin2020dense}:
(1) \textbf{Random Noise}: Returns randomly selected passages for a given query.
(2) \textbf{BM25} \cite{robertson2009probabilistic}: A sparse retrieval method based on exact token matching and term frequency.
(3) \textbf{E5 (HNSW)} \cite{malkov2018efficient,wang2022text}: A dense retrieval method that encodes queries and passages into semantic embeddings, using dot product similarity for matching. HNSW provides efficient approximate nearest neighbor (ANN) search at the cost of some accuracy.
(4) \textbf{E5 (Exact Match)} \cite{wang2022text}: A dense retrieval method using exact embedding matching without approximation, ensuring the highest retrieval accuracy.
The retrieval performance of these methods follows the ranking: E5 (Exact) > E5 (HNSW) > BM25 > Random, as shown in Figure~\ref{fig:retriever-training-study}(a).
Detailed experimental settings and case studies can be found in Appendix \ref{apx:setting5} and \ref{apx:sec:train-engine}.

\textbf{Results.}
The training reward curves and final test performance under different search engine settings are presented in Figure~\ref{fig:retriever-training-study}(c) and Table~\ref{tab:train-retriever}. We observe the following trends:
(1) Training with stronger search engines (\textit{e.g.}, \textbf{E5 (Exact)} and \textbf{E5 (HNSW)}) results in more stable RL training and better final performance.
(2) Training with weaker search engines (\textit{e.g.}, \textbf{Random} and \textbf{BM25}) leads to suboptimal final performance.
The search engine call frequency during training is illustrated in Figure~\ref{fig:retriever-training-study}(d), revealing:
(1) With \textbf{Random Noise}, the agent quickly learns to avoid using the search engine, as the retrieved information does not contribute to problem-solving.
(2) With \textbf{BM25}, the agent gradually increases the number of search engine calls. Since BM25 offers limited retrieval quality, the agent needs to issue multiple, refined queries to obtain relevant information.
(3) With \textbf{E5} (a stronger retriever), the agent learns to utilize the search engine judiciously, making a reasonable number of calls to acquire the necessary information efficiently.

\subsection{Inference with Different Search Engines.}\label{sec:infer-engine}

\begin{table}[t]
    \centering
    \scriptsize
    \setlength{\tabcolsep}{1pt}
    \renewcommand{\arraystretch}{1.4}
    \caption{Retriever generalization results across datasets and test retrievers. (Qwen2.5-7B-Base, PPO)}\label{tab:retriever-generalization}
    \begin{tabular}{lcccccccccccc}
        \toprule
        \textbf{Train / Test Search Engine} & \multicolumn{3}{c}{\textbf{BM25}} & \multicolumn{3}{c}{\textbf{E5 (HNSW)}} & \multicolumn{3}{c}{\textbf{E5 (Exact)}} & \multicolumn{3}{c}{\textbf{Google Search}} \\
        \cmidrule(lr){2-4} \cmidrule(lr){5-7} \cmidrule(lr){8-10} \cmidrule(lr){11-13}
         & \textbf{Bambg} & \textbf{GPQA} & \textbf{SimpleQA} & \textbf{Bambg} & \textbf{GPQA} & \textbf{SimpleQA} & \textbf{Bambg} & \textbf{GPQA} & \textbf{SimpleQA} & \textbf{Bambg} & \textbf{GPQA} & \textbf{SimpleQA} \\
        \midrule
        BM25 & 0.280 & 0.273 & 0.243 & 0.432 & 0.293 & 0.159 & 0.424 & 0.323 & 0.259 & 0.496 & 0.313 & 0.540 \\
        E5 (HNSW) & 0.240 & 0.298 & 0.270 & 0.400 & 0.288 & 0.169 & 0.440 & 0.273 & 0.254 & 0.528 & 0.333 & 0.603 \\
        E5 (Exact) & 0.312 & 0.313 & 0.249 & 0.400 & 0.298 & 0.196 & 0.424 & 0.288 & 0.265 & 0.560 & 0.293 & 0.603 \\
        \midrule
        Average & 0.277 & 0.295 & 0.254 & 0.411 & 0.293 & 0.175 & 0.429 & 0.295 & 0.259 & 0.528 & 0.313 & 0.582   \\
        \bottomrule
    \end{tabular}
\end{table}

\textbf{Motivation.}
In practical scenarios, certain search engines may be unsuitable for integration during RL training due to empirical constraints such as accessibility, cost, or API limitations. 
In such cases, it becomes necessary to train the agent using one search engine while employing a different one during inference. 
This raises important research questions: (1) To what extent does the difference in search engines between training and inference affect model performance? (2) Does utilizing a stronger search engine at inference lead to improved downstream performance?

\textbf{Experimental Design.}
Following the training setup in \cite{jin2025search}, we investigate these questions by training the LLM-based search agent using three different search engines: 
(1) BM25 \cite{robertson2009probabilistic}, 
(2) E5 \cite{wang2022text} with approximate nearest neighbor (ANN) search implemented via HNSW \cite{malkov2018efficient}, and 
(3) E5 with exact search. 
During inference, we additionally include the online Google Search API\footnote{\url{https://developers.google.com/custom-search/v1/overview}} as a stronger retrieval system. 
To study the importance of the search engine in challenging scenarios, we construct a combined benchmark consisting of 512 samples. This benchmark includes the full Bamboogle test set~\cite{press2022measuring} (Bambg), the complete GPQA-diamond dataset~\cite{rein2024gpqa} (GPQA), and a randomly sampled subset from SimpleQA~\cite{wei2024measuring}.
Detailed experimental settings can be found in Appendix \ref{apx:setting6}.

\textbf{Results.}
As shown in Table \ref{tab:retriever-generalization}, we observe the following:
(1) LLM search agents trained with a specific search engine demonstrate strong generalization capabilities when evaluated with different search engines during inference.
(2) Leveraging a more powerful search engine at inference time (\textit{e.g.}, Google Search) consistently and significantly leads to improved performance, highlighting the importance of high-quality retrieval in downstream applications.

More studies on long-form generation tasks with outcome drive RL and data scaling study can be found in Appendix \ref{apx:sec:longform} and \ref{apx:sec:data-scaling}, respectively.

\section{Conclusion}
In this work, we conduct comprehensive empirical studies on key design factors in training LLM-based search agents using reinforcement learning. 
Our investigation reveals that format rewards play an important role in certain scenarios, while intermediate retrieval rewards provide limited benefit and may not consistently improve the learning process.
We demonstrate that the choice of the underlying LLM (whether a general-purpose model or one specialized for reasoning) and its scale significantly affect the final agent’s performance. 
Additionally, the selection of the search engine plays a non-trivial role in shaping both the RL training dynamics and the robustness of the agent during inference. 
These insights offer practical guidance for developing more capable and reliable LLM-based search agents, paving the way for their deployment in real-world applications.
Interesting future directions include exploring more advanced reward modeling techniques, such as learned reward functions and preference-based feedback, as well as studying the agentic behaviors acquired through RL in broader scenarios, including tool use and software engineering.

\begin{ack}
This research was supported in part by Apple PhD Fellowship, in part by US DARPA INCAS Program No. HR0011-21-C0165 and BRIES Program No. HR0011-24-3-0325, in part by the Office of Naval Research contract number N000142412612, in part by NSF grant numbers IIS-19-56151 and 2402873, in part by the Molecule Maker Lab Institute: An AI Research Institutes program supported by NSF under Award No. 2019897 and the Institute for Geospatial Understanding through an Integrative Discovery Environment (I-GUIDE) by NSF under Award No. 2118329, in part by Cisco, and in part by the Center for Intelligent Information Retrieval. Any opinions, findings, and conclusions or recommendations expressed herein are those of the authors and do not necessarily represent the views, either expressed or implied, of the sponsors or the U.S. Government.
\end{ack}

\bibliographystyle{plainnat}
\bibliography{references}


\newpage
\appendix

\section{Limitations}\label{apx:sec:limitation}

In this work, we conduct an empirical investigation into the use of reinforcement learning (RL) for training LLM-based search agents. 
Our study focuses on three key factors that influence the effectiveness of RL in this context: (1) reward formulation, (2) the choice and characteristics of the underlying language model, and (3) the role and quality of the search engine.
However, our analysis is primarily confined to search-based agent scenarios. 
While this provides valuable insights into the challenges and design choices for RL in retrieval-augmented reasoning, it does not fully generalize to other classes of LLM-based agents. 
In particular, RL for more complex and open-ended agent behaviors---such as those exhibited by data science assistants, software engineering agents, or multi-tool task planners—remains underexplored. 
These domains may require more sophisticated reward structures, long-horizon credit assignment, and multi-step decision-making policies that go beyond the scope of search-oriented tasks.
We leave a more comprehensive study of RL in broader agentic settings as an important future work direction.

\section{Positive and Negative Societal Impacts}\label{apx:sec-impact}

Our work aims to improve the efficiency and reliability of large language model (LLM)-based agents by leveraging reinforcement learning to optimize their interaction with external tools such as search engines. 
On the positive side, this line of research can enhance the capabilities of LLMs in high-stakes domains like scientific research, education, and healthcare, where accurate information retrieval and reasoning are critical. 
By making LLM-based agents more effective at querying and utilizing external knowledge sources, our approach has the potential to reduce hallucinations, improve transparency, and increase user trust in AI systems.
However, as with all advances in powerful LLM-based agents, there are potential negative impacts. 
Improved autonomy in tool usage may lead to unintended misuse, such as generating convincing but misleading information or automating complex tasks without sufficient human oversight.
Moreover, the deployment of search-augmented agents could exacerbate access disparities if such technologies are restricted to proprietary systems.
Careful consideration of ethical deployment, transparency in agent decision-making, and equitable access to advanced AI capabilities is essential to mitigate these risks.

\section{Format Reward Code}\label{apx:sec:format}

In this section, we provide the code to judge whether the rollout sequence is in a desired format for a search agent.

\begin{lstlisting}[language=Python, caption=Format Reward Code., label={lst:example_python}, 
    basicstyle=\ttfamily\small, 
    keywordstyle=\color{blue}, 
    stringstyle=\color{red}, 
    commentstyle=\color{gray}, 
    showstringspaces=false, 
    breaklines=true]

def is_valid_sequence(text):
    # Find the position of "<|im_start|>assistant" with potential whitespace
    assistant_pattern = r"<\|im_start\|>assistant\s*"
    assistant_match = re.search(assistant_pattern, text)
    
    if not assistant_match:
        return False, "Missing assistant marker"
    
    # Extract the content after the assistant marker
    start_pos = assistant_match.end()
    content = text[start_pos:]
    
    # Check for balanced tags
    tags_to_check = ["think", "search", "information", "answer"]
    for tag in tags_to_check:
        opening_count = len(re.findall(f"<{tag}>", content))
        closing_count = len(re.findall(f"</{tag}>", content))
        if opening_count != closing_count:
            return False, f"Mismatch in {tag} tags: {opening_count} opening vs {closing_count} closing tags"
    
    # Now check for proper sequence pattern and no extraneous content
    
    # 1. First split the content by any tags we recognize
    split_pattern = r"(</?(?:think|search|information|answer)>)"
    parts = re.split(split_pattern, content)
    
    # 2. Keep track of the current position in the expected sequence
    state = "start"  # start -> think -> search -> information -> think -> ... -> answer -> end
    
    # 3. Check each part
    for i, part in enumerate(parts):
        # Skip empty parts
        if not part.strip():
            continue
            
        # Check if this is a tag
        if re.match(r"</?(?:think|search|information|answer)>", part):
            # This is a tag, check if it's valid in the current state
            if part == "<think>" and state in ["start", "information"]:
                state = "in_think"
            elif part == "</think>" and state == "in_think":
                state = "after_think"
            elif part == "<search>" and state == "after_think":
                state = "in_search"
            elif part == "</search>" and state == "in_search":
                state = "after_search"
            elif part == "<information>" and state == "after_search":
                state = "in_information"
            elif part == "</information>" and state == "in_information":
                state = "information"
            elif part == "<answer>" and state == "after_think":
                state = "in_answer"
            elif part == "</answer>" and state == "in_answer":
                state = "end"
            else:
                return False, f"Unexpected tag {part} in state {state}"
        else:
            # This is content, check if it's valid in the current state
            if state in ["in_think", "in_search", "in_information", "in_answer"]:
                # Content is allowed inside tags
                pass
            elif state in ["start", "after_think", "after_search", "information"]:
                # Only whitespace is allowed between tags
                if part.strip():
                    return False, f"Unexpected content '{part.strip()}' between tags (state: {state})"
            else:
                return False, f"Unexpected content in state {state}"
    
    # Check final state
    if state != "end":
        return False, f"Incomplete sequence, ended in state {state}"
        
    return True, "Valid sequence format"

\end{lstlisting}

\section{Reward for Long-form Answers}\label{apx:sec:longform}

\textbf{Motivation.}
It is demonstrated that rule-based outcome rewards are effective for training LLM-based search agents~\citep{jin2025search,zheng2025deepresearcher}.
However, their evaluation primarily focuses on short-form QA tasks, where answer correctness can be reliably measured using exact string matching.
In real-world applications, many queries require long-form, open-ended answers, where evaluating correctness is inherently more subjective.
We investigate whether rule-based outcome rewards remain effective in training LLM search agents for long-form QA tasks.

\textbf{Experimental Design.}
We conduct experiments on two long-form QA datasets: ASQA \cite{stelmakh2022asqa} and ELI5 \cite{fan2019eli5}.
Models are trained on the ASQA training set and evaluated on its development set for in-distribution performance.
Out-of-distribution performance is evaluated on the ELI5 dataset.
Following common practice, we use the F1 score as the rule-based evaluation metric for both training and evaluation.
Experiments are conducted on both Qwen2.5-3B-Base and Qwen2.5-7B-Base models.
We compare against several baselines, including Direct Inference (with instruct LLMs), RAG \cite{gao2023retrieval} (with instruct LLMs), and R1 \cite{guo2025deepseek}.
Notably, R1 represents an LLM, trained using the RL approach from DeepSeek-R1 with the same training data (ASQA training set).
We also explore a variant of ~\citep{jin2025search,zheng2025deepresearcher} that incorporates the format reward, denoted as ``\Ours w. Outcome + Format reward''.

\begin{table}[h]
    \centering
    \scriptsize
    \setlength{\tabcolsep}{4pt}
    \renewcommand{\arraystretch}{1.2}
    \caption{Study on long-form question answering tasks.}\label{tab:long-form}
    \begin{tabular}{lccccccccc}
        \toprule
        \textbf{Methods} & \multicolumn{3}{c}{\textbf{Qwen2.5-3b}} & \multicolumn{3}{c}{\textbf{Qwen2.5-7b}} & \multicolumn{3}{c}{\textbf{Qwen2.5-14b}} \\
        \cmidrule(lr){2-4} \cmidrule(lr){5-7} \cmidrule(lr){8-10}
         & \textbf{ASQA} & \textbf{ELI5} & \textbf{Avg.} & \textbf{ASQA} & \textbf{ELI5} & \textbf{Avg.} & \textbf{ASQA} & \textbf{ELI5} & \textbf{Avg.} \\
        \midrule
        Direct & 0.251 & 0.199 & 0.225 &  0.303 & 0.201 & 0.252 & 0.289 & 0.199 & 0.244  \\
        RAG & 0.301 & 0.193 & 0.247 & 0.317 & 0.202 & 0.259 & 0.285 & 0.193 & 0.239 \\
        R1 & 0.424 & \textbf{0.275} & 0.350 & 0.437 & \textbf{0.280} &	0.358 & 0.444 & \textbf{0.278} & 0.361  \\
        \hdashline
        \Ours w. Outcome + Format reward (PPO) & 0.480 & 0.261 & 0.370 &  0.471 & 0.256 & 0.363 & 0.442 & 0.260 & 0.351 \\
        \Ours w. Outcome + Format reward (GRPO) & \textbf{0.492} & 0.272 & \textbf{0.382} & \textbf{0.504} & 0.275 & \textbf{0.390} & \textbf{0.501} & 0.273 & \textbf{0.387} \\
        \bottomrule
    \end{tabular}
\end{table}

\textbf{Results.}
Performance comparisons are shown in Table~\ref{tab:long-form}.
We observe that ``\Ours w. Outcome + Format reward'' achieves competitive results on long-form QA tasks, demonstrating the effectiveness of rule-based outcome rewards with format rewards even in complex, open-ended scenarios.
This suggests that RL guided by the proposed rewards remains a viable strategy for training search-augmented LLMs, even when the evaluation objective shifts from short-form to long-form answer generation.

\section{Study of Data Scaling}\label{apx:sec:data-scaling}

\textbf{Motivation.}
Although RL has shown strong potential in training LLM-based search agents \cite{jin2025search}, the impact of training data size on the RL process remains underexplored.
While recent work has shown that extremely small datasets can be sufficient for reasoning-oriented RL with LLMs \cite{wang2025reinforcement}, it is unclear whether similar data efficiency holds in \textit{agentic RL} settings, where models must learn to reason and interact with external tools in an interleaved manner.

\textbf{Experimental Design.}
Following \cite{jin2025search}, we use the training sets from NQ \cite{kwiatkowski2019natural} and HotpotQA \cite{yang2018hotpotqa} as our full training set.
To investigate the effect of training data size, we construct subsets of varying sizes by randomly sampling $k$ examples from the full dataset, where $k \in {1, 10, 100, 1000, 10000}$.
We conduct experiments using the Qwen2.5-3B-Base model and evaluate both PPO and GRPO as the underlying reinforcement learning algorithms.

\begin{table}[h]
    \centering
    \scriptsize
    \setlength{\tabcolsep}{4pt}
    \renewcommand{\arraystretch}{1.2}
    \caption{Final performance with different size of training data. $\mathcal{D}$ is the training data. The best performance is set in bold. $^\dagger/^\star$ represents in-domain/out-domain datasets. (LLM: Qwen2.5-3B-Base)}\label{apx:tab:data-scaling}
    \begin{tabular}{lcccccccc}
        \toprule
        \textbf{Methods} & \multicolumn{3}{c}{\textbf{General QA}} & \multicolumn{4}{c}{\textbf{Multi-Hop QA}} \\
        \cmidrule(lr){2-4} \cmidrule(lr){5-8}
         & \textbf{NQ$^\dagger$} & \textbf{TriviaQA$^\star$} & \textbf{PopQA$^\star$} & \textbf{HotpotQA$^\dagger$} & \textbf{2wiki$^\star$} & \textbf{Musique$^\star$} & \textbf{Bamboogle$^\star$} & \textbf{Avg.} \\
        \midrule
        \multicolumn{9}{l}{\textbf{PPO}} \\
        \hdashline
        $|\mathcal{D}|$ = 1 & 0.121 & 0.339 & 0.119 & 	0.141 & 0.211 & 0.017 & 0.056 & 0.143  \\
        $|\mathcal{D}|$ = 10 & 0.339 & 0.509 & 0.379 & 0.236 & 0.237 & 0.048 & 0.081	& 0.261  \\
        $|\mathcal{D}|$ = 100 & 0.372 & 0.549 & 0.382 & 0.262 & 0.279 & 0.063 & 0.161 & 0.295  \\
        $|\mathcal{D}|$ = 1000 & \textbf{0.431} & \textbf{0.599} & \textbf{0.446} & 0.348 & 0.355 & 0.136 & 0.298 & 0.373  \\
        $|\mathcal{D}|$ = 10000 &  0.430 & 0.594 & 0.445 & \textbf{0.369} & \textbf{0.383} & \textbf{0.155} & \textbf{0.315} & \textbf{0.384} \\
        \midrule
        \multicolumn{9}{l}{\textbf{GRPO}} \\
        \hdashline
        $|\mathcal{D}|$ = 1 & 0.107 & 0.287 & 0.115 & 0.124 & 0.202 & 0.019 & 0.089 & 0.134  \\
        $|\mathcal{D}|$ = 10 & 0.338 & 0.515 & 0.361 & 0.236 & 0.223 & 0.043 & 0.089 & 0.258  \\
        $|\mathcal{D}|$ = 100 & 0.367 & 0.529 & 0.414 & 0.265 & 0.302 & 0.079 & 0.194 & 0.307  \\
        $|\mathcal{D}|$ = 1000 & 0.421 & 0.594 & \textbf{0.437} & 0.363 & 0.364 & \textbf{0.149} & \textbf{0.315} & 0.377  \\
        $|\mathcal{D}|$ = 10000 &  \textbf{0.435} & \textbf{0.599} & 0.435 & \textbf{0.365} & \textbf{0.379} & 0.137 & 0.306 & \textbf{0.379} \\
        \bottomrule
    \end{tabular}
\end{table}

\begin{figure}[h]
    \centering
    \subfigure[Training Reward]{%
        \includegraphics[width=0.23\textwidth]{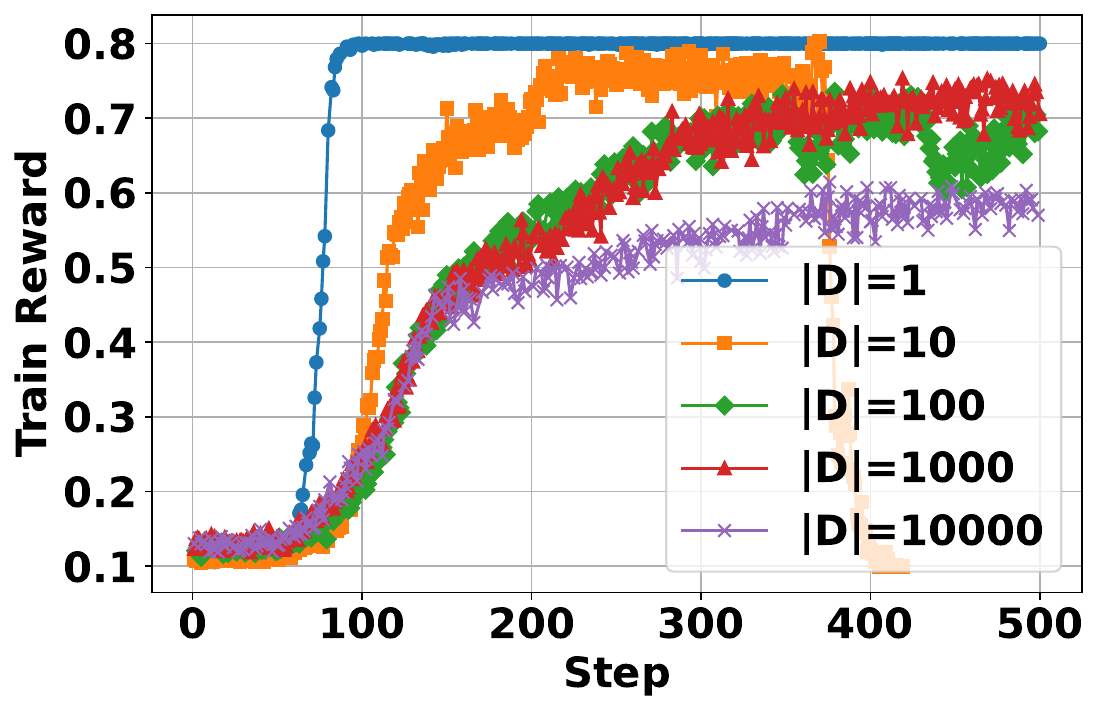}
    }
    \subfigure[\# of Search Calls]{%
        \includegraphics[width=0.23\textwidth]{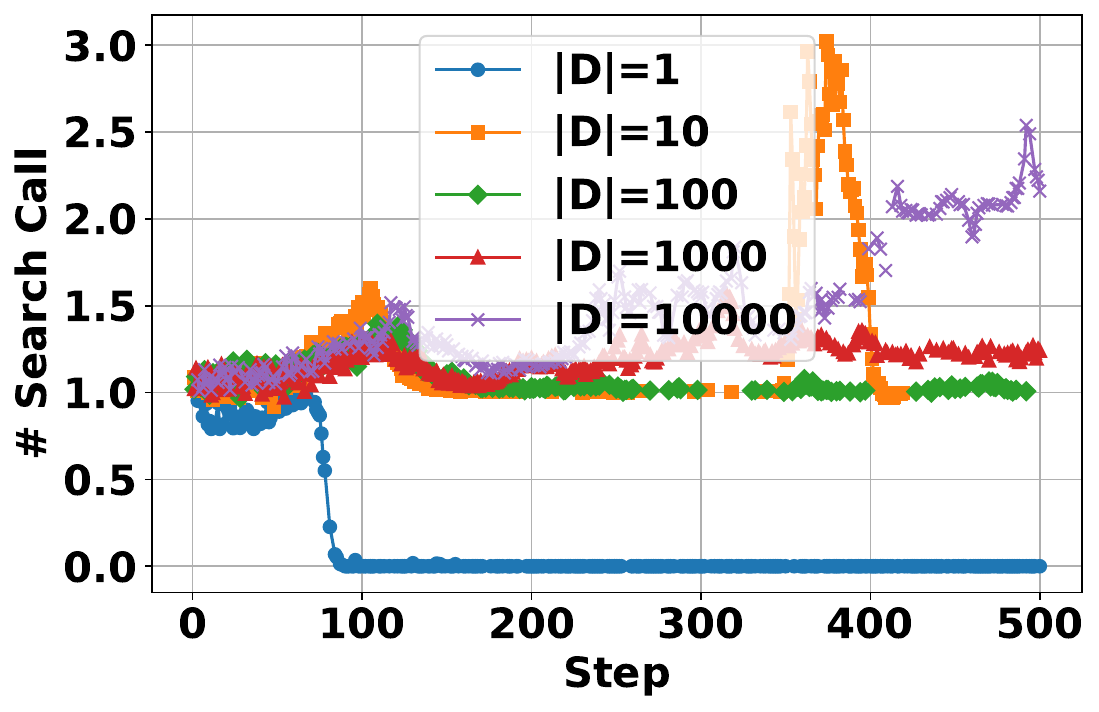}
    }
    \subfigure[Training Reward]{%
        \includegraphics[width=0.24\textwidth]{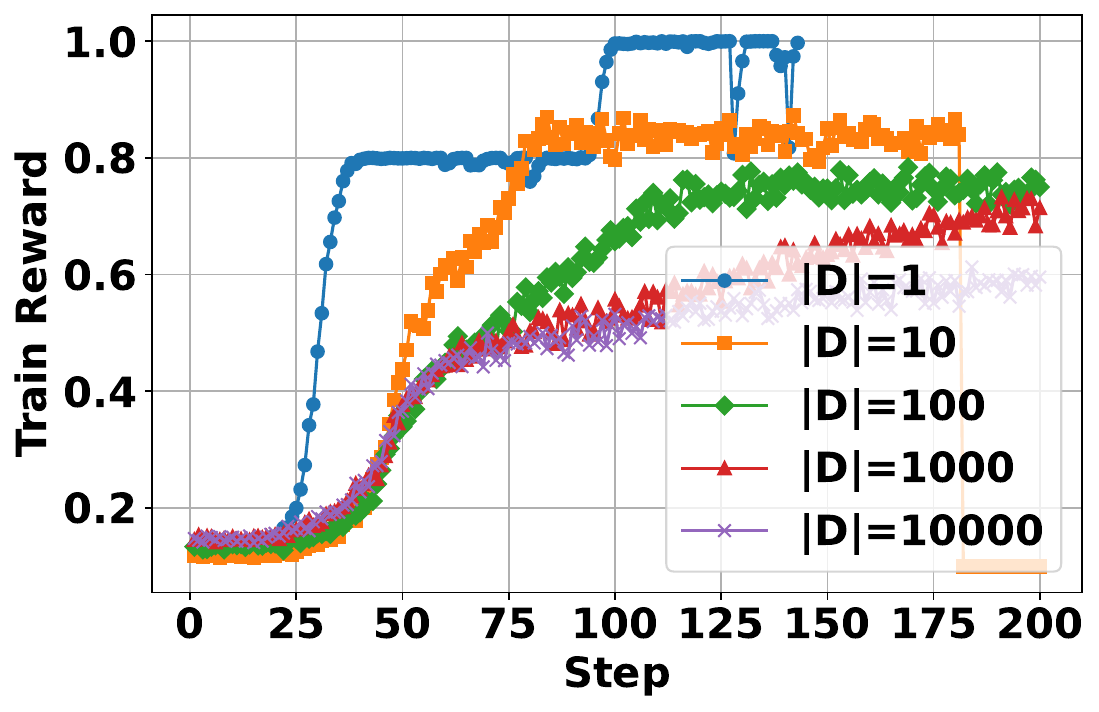}
    }
    \subfigure[\# of Search Calls]{%
        \includegraphics[width=0.24\textwidth]{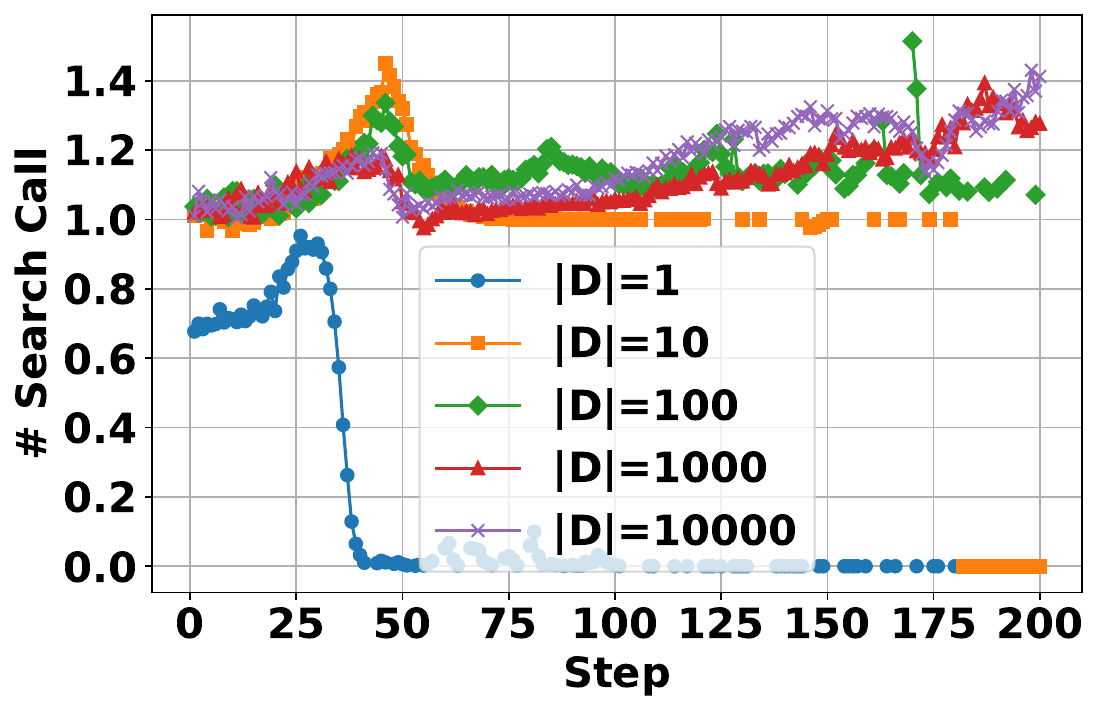}
    }
    \caption{
Data scaling effects in RL training for search agents.
(a) \textbf{Training reward under PPO with varying dataset sizes}: Smaller training sets result in faster convergence and higher training rewards, likely due to overfitting.
(b) \textbf{Number of search engine calls under PPO}: Training with a single example fails to induce search behavior, while 10 samples lead to unstable learning. In contrast, using 100 or 1,000 samples enables the model to learn stable search behavior, and training with 10,000 samples further improves performance.
(c) \textbf{Training reward under GRPO with varying dataset sizes}: Similar to PPO, smaller datasets yield faster convergence and higher rewards, again suggesting potential overfitting.
(d) \textbf{Number of search engine calls under GRPO}: A single training sample is insufficient for search behavior to emerge, whereas larger datasets facilitate stable learning of search interactions.
}\label{apx:fig:scaling-data}
\end{figure}

\textbf{Results.}
We present the training reward dynamics and the number of search engine calls across varying training data sizes in Figure~\ref{apx:fig:scaling-data}. The final performance of LLM-based search agents trained with different dataset sizes is reported in Table~\ref{apx:tab:data-scaling}. The results reveal several key observations:
(1) Increasing the size of the training dataset generally leads to improved performance, particularly on more complex multi-hop question answering tasks such as HotpotQA and 2Wiki.
(2) Smaller datasets lead to faster convergence and higher training rewards, which is likely attributable to overfitting.
(3) For PPO, training with a single example fails to induce meaningful search behavior, while using 10 examples results in unstable training. In contrast, training with 100 or 1,000 examples enables the model to learn stable search behavior, and performance continues to improve with 10,000 examples. Similar trends are observed under GRPO training.

\section{More Studies on LLM Types}\label{apx:sec:llm-type}

\begin{figure}[h]
    \centering
    \subfigure[Train Reward]{%
        \includegraphics[width=0.4\textwidth]{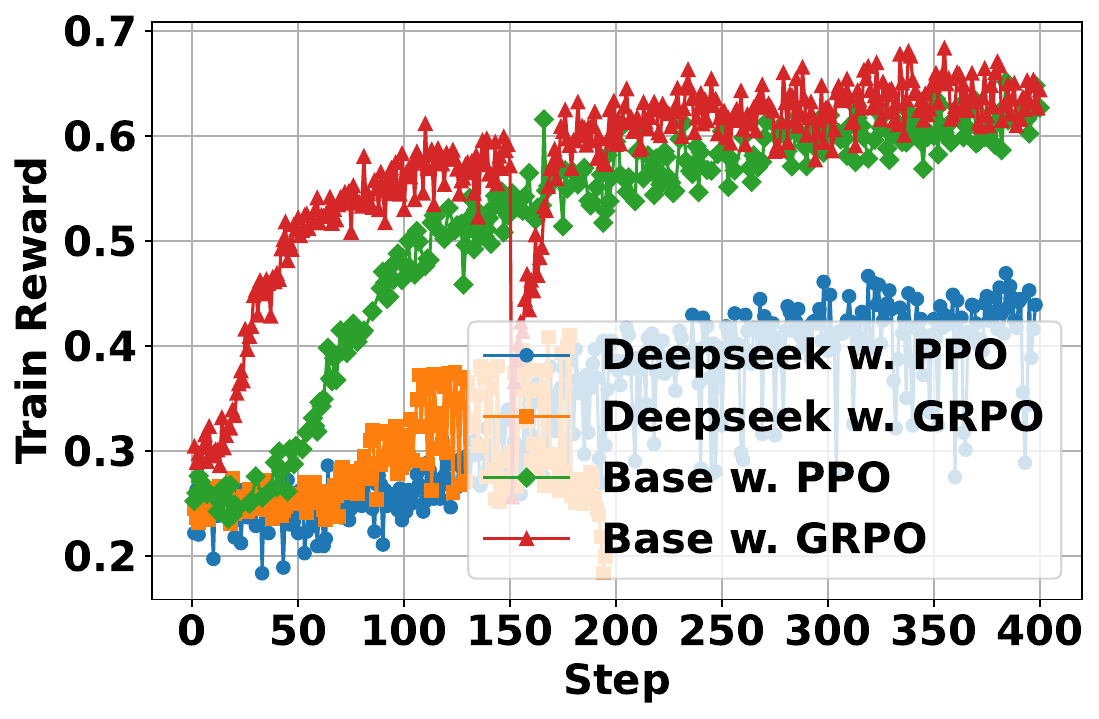}
    }
    \hspace{0.1in}
    \subfigure[\# of Search Calls]{%
        \includegraphics[width=0.4\textwidth]{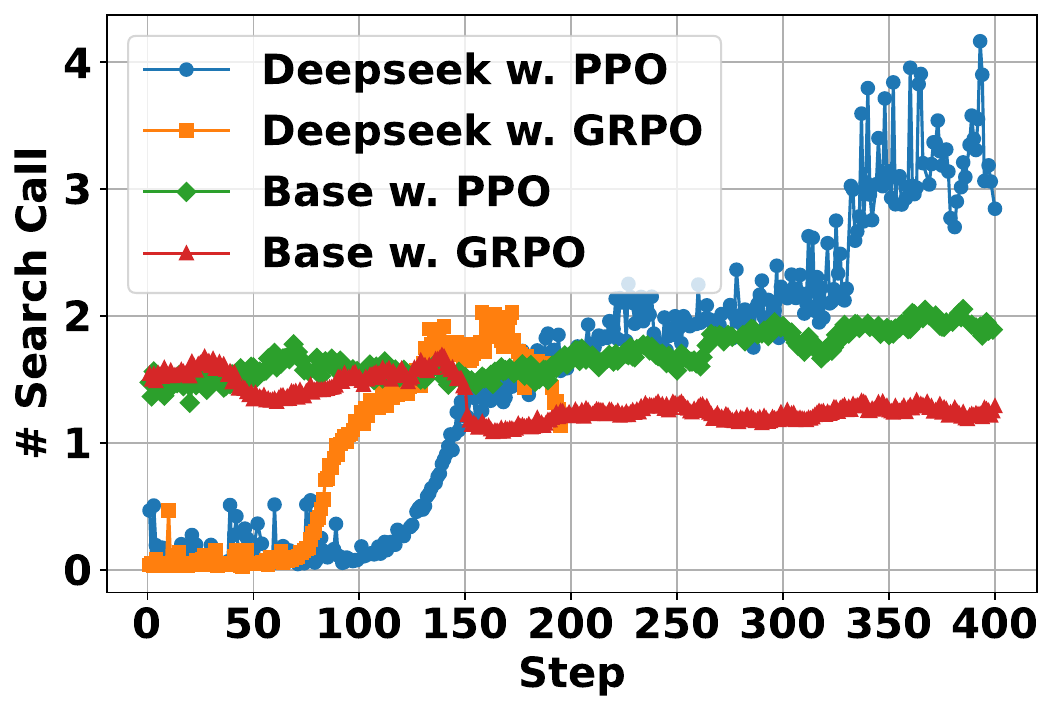}
    }
    \caption{
The study of underlying pretrained LLM for development of search agents with RL. 
(a) \textbf{Training reward with different type of LLMs} - general-purpose LLM (Qwen2.5-14B-Base) and reasoning LLM (DeepSeek-R1-Distill-Qwen-14B). We observe that general-purpose LLM performs better than reasoning LLMs with both PPO and GRPO.
(b) \textbf{\# of Search engine calls with different type of LLMs}: Both the general-purpose LLM and the reasoning-specialized LLM demonstrate the ability to learn when to call the search engine. However, the general-purpose LLM achieves better final performance, potentially due to its superior capability in formulating effective search queries.
}\label{apx:fig:scaling-LLM}
\end{figure}

In addition to the 7B model analysis in Section~\ref{sec:llm-type}, we further investigate the impact of LLM initialization on RL performance using 14B-scale models. Specifically, we compare Qwen2.5-14B-Base as a general-purpose LLM and DeepSeek-R1-Distill-Qwen-14B as a reasoning-specialized LLM, evaluating both under PPO and GRPO training. The results, summarized in Table~\ref{apx:tab:llm-type} and Figure~\ref{apx:fig:scaling-LLM}, yield the following observations:
(1) RL training is more stable and effective when initialized from the general-purpose LLM, suggesting that such models possess sufficient general reasoning capabilities to support the search agent task, even without reasoning-specific pretraining.
(2) Although the reasoning-specialized LLM eventually learns to perform interleaved reasoning and retrieval with PPO and GRPO, the general-purpose LLM consistently achieves higher final performance—likely due to its stronger ability to generate effective search queries.

\begin{table}[h]
    \centering
    \scriptsize
    \setlength{\tabcolsep}{4pt}
    \renewcommand{\arraystretch}{1.2}
    \caption{LLM type study with 14B LLMs. The best performance is set in bold. $^\dagger/^\star$ represents in-domain/out-domain datasets.}\label{apx:tab:llm-type}
    \begin{tabular}{lcccccccc}
        \toprule
        \textbf{Methods} & \multicolumn{3}{c}{\textbf{General QA}} & \multicolumn{4}{c}{\textbf{Multi-Hop QA}} \\
        \cmidrule(lr){2-4} \cmidrule(lr){5-8}
         & \textbf{NQ$^\dagger$} & \textbf{TriviaQA$^\star$} & \textbf{PopQA$^\star$} & \textbf{HotpotQA$^\dagger$} & \textbf{2wiki$^\star$} & \textbf{Musique$^\star$} & \textbf{Bamboogle$^\star$} & \textbf{Avg.} \\
        \midrule
        \multicolumn{9}{l}{\textbf{DeepSeek-R1-Distill-Qwen-14b}} \\
        \hdashline
        PPO & 0.475 & 0.634 & 0.465 & 0.401 & 0.363 & 0.211 & 0.476	& 0.432  \\
        GRPO & 0.305 & 0.613 & 0.332 & 0.285 & 0.276 & 0.092 & 0.347 & 0.321  \\
        \midrule
        \multicolumn{9}{l}{\textbf{Qwen2.5-14b-Base}} \\
        \hdashline
        PPO & 0.499 & 0.680 & 0.472 & 0.452 & 0.431 & 0.215 & 0.468 & 0.459 \\
        GRPO &  \textbf{0.500} & \textbf{0.693} & \textbf{0.500} & \textbf{0.481} & \textbf{0.488} & \textbf{0.261} & \textbf{0.516} & \textbf{0.491}  \\
        \bottomrule
    \end{tabular}
\end{table}

\newpage
\section{Experimental Settings}\label{apx:sec:setting}

In this section, we detail the experimental settings used in the studies presented in Sections~\ref{sec:reward-study}, \ref{sec:backbone-llm}, \ref{sec:search-engine}.

\subsection{Section \ref{sec:format-reward}}\label{apx:setting1}

We adopt the same training dataset as used in \cite{jin2025search}, consisting of the Natural Questions (NQ) and HotpotQA training sets.
For PPO training, the policy LLM learning rates are set to $1 \times 10^{-6}$ for Qwen2.5-3B and Qwen2.5-7B, and $5 \times 10^{-7}$ for Qwen2.5-14B. The critic LLM learning rate is fixed at $1 \times 10^{-5}$ across all model sizes.
For GRPO training, we use a policy LLM learning rate of $5 \times 10^{-7}$ for all models. The RL training batch size is set to 512, and the rollout temperature is fixed at 1. For GRPO, we set the group size to 5.
Each model is trained for up to 600 steps, with early stopping triggered if training collapse is observed based on the reward curve. For the results reported in Table~\ref{tab:format}, we use either the final checkpoint at step 600 or the last checkpoint prior to collapse.
All training jobs are conducted on a node equipped with 8 NVIDIA H100 GPUs.
We conduct a grid search over $\lambda_f \in {0.2, 0.4, 0.6, 0.8}$ and select the best-performing value for each model: 0.2 for 3B, 0.4 for 7B, and 0.2 for 14B.
We adopt E5 (exact) as the retriever and return the top-3 passages (each contains about 200 tokens).

\subsection{Section \ref{sec:retrieval-reward}}\label{apx:setting2}

We adopt the same training dataset as used in \cite{jin2025search}, consisting of the Natural Questions (NQ) and HotpotQA training sets.
For PPO training, the policy LLM learning rates are set to $1 \times 10^{-6}$ for both Qwen2.5-3B and Qwen2.5-7B. The critic LLM learning rate is fixed at $1 \times 10^{-5}$ across all model sizes.
For GRPO training, we use a policy LLM learning rate of $5 \times 10^{-7}$ for both models. The RL training batch size is set to 512, and the rollout temperature is fixed at 1. For GRPO, we set the group size to 5.
Each model is trained for up to 600 steps, with early stopping triggered if training collapse is observed based on the reward curve. For the results reported in Table~\ref{tab:format}, we use either the final checkpoint at step 600 or the last checkpoint prior to collapse.
All training jobs are conducted on a node equipped with 8 NVIDIA H100 GPUs.
Based on the findings in Section~\ref{sec:format-reward}, we fix $\lambda_f$ to 0.2 for the 3B model and 0.4 for the 7B model. 
We then perform a grid search over $\lambda_r \in {0.1, 0.3, 0.5}$ and select the best-performing value for each model, which is 0.1 for both 3B and 7B.
We adopt E5 (exact) as the retriever and return the top-3 passages (each contains about 200 tokens).

\subsection{Section \ref{sec:llm-type}}\label{apx:setting3}

We adopt the same training dataset as used in \cite{jin2025search}, consisting of the Natural Questions (NQ) and HotpotQA training sets.
For PPO training, the policy LLM learning rates are set to $1 \times 10^{-6}$ for both Qwen2.5-7B and DeepSeek-R1-Distill-Qwen-7B. The critic LLM learning rate is fixed at $1 \times 10^{-5}$ across all model sizes.
For GRPO training, we use a policy LLM learning rate of $5 \times 10^{-7}$ for both models. The RL training batch size is set to 512, and the rollout temperature is fixed at 1. For GRPO, we set the group size to 5.
Each model is trained for up to 600 steps, with early stopping triggered if training collapse is observed based on the reward curve. For the results reported in Table~\ref{tab:format}, we use either the final checkpoint at step 600 or the last checkpoint prior to collapse.
All training jobs are conducted on a node equipped with 8 NVIDIA H100 GPUs.
We set $\lambda_f$ as 0.2 and $\lambda_r$ as 0 for all the experiment.
We adopt E5 (exact) as the retriever and return the top-3 passages (each contains about 200 tokens).

\subsection{Section \ref{sec:llm-scale-up}}\label{apx:setting4}

We adopt the same training dataset as used in \cite{jin2025search}, consisting of the Natural Questions (NQ) and HotpotQA training sets.
For GRPO training, we use a policy LLM learning rate of $5 \times 10^{-7}$ for all models. The RL training batch size is set to 512, and the rollout temperature is fixed at 1. For GRPO, we set the group size to 5.
Each model is trained for up to 600 steps, with early stopping triggered if training collapse is observed based on the reward curve. For the results reported in Table~\ref{tab:format}, we use either the final checkpoint at step 600 or the last checkpoint prior to collapse.
All training jobs are conducted on a node equipped with 8 NVIDIA H100 GPUs.
We set $\lambda_f$ as 0.2 for all the experiment.
We adopt E5 (exact) as the retriever and return the top-3 passages (each contains about 200 tokens).

\subsection{Section \ref{sec:train-engine}}\label{apx:setting5}

We adopt the same training dataset as used in \cite{jin2025search}, consisting of the Natural Questions (NQ) and HotpotQA training sets.
We adopt Qwen2.5-7B-Base as the LLM backbone and PPO as the RL method.
For PPO training, the policy LLM learning rates are set to $1 \times 10^{-6}$ and the critic LLM learning rate is fixed at $1 \times 10^{-5}$.
Each model is trained for up to 600 steps, with early stopping triggered if training collapse is observed based on the reward curve. 
For the results reported in Table~\ref{tab:format}, we use either the final checkpoint at step 600 or the last checkpoint prior to collapse.
All training jobs are conducted on a node equipped with 8 NVIDIA H100 GPUs.
For BM25, we adopt the Pyserini implementation \footnote{https://github.com/castorini/pyserini} and for E5, we adopt Faiss \footnote{https://github.com/facebookresearch/faiss}.
We adopt ``HNSW64'' for the ANN setting.
We return the top-3 passages (each contains about 200 tokens) from each search engine.

\subsection{Section \ref{sec:infer-engine}}\label{apx:setting6}

We adopt the same training dataset as used in \cite{jin2025search}, consisting of the Natural Questions (NQ) and HotpotQA training sets.
We adopt Qwen2.5-7B-Base as the LLM backbone and PPO as the RL method.
For PPO training, the policy LLM learning rates are set to $1 \times 10^{-6}$ and the critic LLM learning rate is fixed at $1 \times 10^{-5}$.
Each model is trained for up to 600 steps, with early stopping triggered if training collapse is observed based on the reward curve. 
For the results reported in Table~\ref{tab:format}, we use either the final checkpoint at step 600 or the last checkpoint prior to collapse.
All training jobs are conducted on a node equipped with 8 NVIDIA H100 GPUs.
For BM25, we adopt the Pyserini implementation and for E5, we adopt Faiss.
We adopt ``HNSW64'' for the ANN setting.
Given that Google Search API is expensive, we construct a testing set with 512 samples, including full test set from Bamboogle \cite{press2022measuring} (125 samples), GPQA-Diamond \cite{rein2024gpqa} (198 samples) and randomly selected 189 samples from SimpleQA \cite{wei2024measuring}.
We return the top-3 passages (each contains about 200 tokens) from each search engine.

\section{Case Studies on LLMs Trained with Different Search Engines}\label{apx:sec:train-engine}

\begin{table}[H]
    \centering
    \footnotesize
    \caption{Case study A of LLM-based search agent trained with different search engines (\textit{i.e.}, random noise, BM25, and E5).}\label{apx:tab:case1}
    \begin{tabular}{p{13.5cm}}
        \hline
        \textbf{Question}: Which Hells Angel member stabbed and beat an attendant of the 1969 counterculture-era rock concert in the United States? \\
        \hline
        \textbf{Ground Truth}: Alan Passaro \\
        \hline
        \textbf{w. random noise}: \\
        \think{To answer this question, we need to recall the incident involving the 1969 counterculture-era rock concert known as Woodstock. The incident in question is the killing of festival security guard Thomas Benson. The Hells Angel member responsible for the stabbing and subsequent death of Benson is.CompareTag Stevenson.} \\ 
        \answer{Manson} \\
        \hline
        \textbf{w. BM25}: \\
        \think{To answer this question, I'll need to identify the Hells Angel member who was involved in the stabbing and beating of an attendant at a 1969 counterculture-era rock concert in the United States. I'll use a search engine to gather information about this event and the responsible Hells Angel member.} \\
\search{Hells Angel member at 1969 counterculture-era rock concert in the United States} \\
\info{Doc 1(Title: "Altamont Free Concert") and for the negative portrayal of the Angels in the ""Gimme Shelter"" film. The conspirators reportedly used a boat to approach a residence where Jagger was staying on Long Island, New York, the plot failing when the boat was nearly sunk by a storm. Jagger's spokesperson has refused to comment on the matter ...
} \\
\hdashline
\think{From the search results, I've identified that the 1969 counterculture-era rock concert in question is the Altamont Free Concert. Now, I need to find out which Hells Angel member was involved in stabbing and beating an attendant at this concert.} \\
\search{Hells Angel member who stabbed and beat an attendant at Altamont Free Concert} \\
\info{Doc 1(Title: "The Rolling Stones") and former members of the band. Two species, Petroperla mickjaggeri and Lapisperla keithrichardsi, were placed within a new family Petroperlidae. The new family was named in honour of the Rolling Stones, derived from the Greek ""petra"" that stands for ""stone"". The scientists referred to the fossils as ""Rolling Stoneflies"". The Rolling Stones' first concert was on 12 July 1962 at the Marquee Club in London ...
} \\
\hdashline
\think{I have found that a Hells Angel member was responsible for stabbing and beating an attendant at the 1969 Altamont Free Concert. To identify which specific Hells Angel member is being referred to, I will search for the individual who committed this act.} \\
\search{individual Hells Angel member who stabbed and beat an attendant at Altamont Free Concert} \\
\info{Doc 1(Title: "The Rolling Stones") and former members of the band. Two species, Petroperla mickjaggeri and Lapisperla keithrichardsi, were placed within a new family Petroperlidae. The new family was named in honour of the Rolling Stones, derived from the Greek ""petra"" that stands for ""stone"". The scientists referred to the fossils as ""Rolling Stoneflies"". The Rolling Stones' first concert was on 12 July 1962 at the Marquee Club in London ...
} \\
\hdashline
\think{After searching for the individual Hells Angel member who stabbed and beat an attendant at the Altamont Free Concert, I have identified that the member was "Ronnie Biggs"} \\
\answer{Charlie ISHAM}\\
\hline
\textbf{w. E5}: \\
\think{I need to find out which Hells Angel member stabbed and beat an attendant of the 1969 counterculture-era rock concert in the United States. I'll search for it.} \\
\search{which Hells Angel member stabbed and beat an attendant of the 1969 counterculture-era rock concert in the United States} \\
\info{Doc 1(Title: "Death of Meredith Hunter") Death of Meredith Hunter Meredith Curly Hunter, Jr. (October 24, 1951 – December 6, 1969) was an 18-year-old African-American teen who was killed at the 1969 Altamont Free Concert. During the performance by The Rolling Stones, Hunter approached the stage, and was violently driven off by members of the Hells ...
} \\
\hdashline
\think{I found out that during the 1969 Altamont Free Concert, a fan named Meredith Hunter was stabbed and beaten to death by Hells Angel Alan Passaro. Now I need to provide the answer.} \\
\answer{Alan Passaro}\\
\hline

\end{tabular}
\end{table}

\begin{table}[H]
    \centering
    \footnotesize
    \caption{Case B study of LLM-based search agent trained with different search engines (\textit{i.e.}, random noise, BM25, and E5).}\label{apx:tab:case2}
    \begin{tabular}{p{13.5cm}}
        \hline
        \textbf{Question}: Jacques Mathou starred in Delicatessen, the 1991 French movie in the style of which celebrated American animator? \\
        \hline
        \textbf{Ground Truth}: Terry Gilliam \\
        \hline
        \textbf{w. random noise}: \\
        \think{To answer this question, we need to consider the animated style of Jacques Mathou's film "Delicatessen." The movie is known ... American animator. Therefore, the celebrated American animator that "Delicatessen" is in the style of is Tim Burton.} \\ 
        \answer{Tim Burton} \\
        \hline
        \textbf{w. BM25}: \\
        \think{To find out which celebrated American animator Jacques Mathou starred in the 1991 French movie "Delicatessen," I will search for information about the movie and its connections ...} \\
\search{Delicatessen 1991 French movie celebrated American animator} \\
\info{Doc 1(Title: "Jean-Pierre Jeunet") Jean-Pierre Jeunet Jean-Pierre Jeunet (; born 3 September 1953) is a French film director, producer, and screenwriter. His films are known to mix elements of fantasy, reality and science fiction either to create idealized realities or to give relevance to mundane situations ...
} \\
\hdashline
\think{From the search results, I have gathered that "Delicatessen" is a 1991 French movie directed by Jean-Pierre Jeunet and Marc Caro. The movie is a post-apocalyptic black comedy. To find the ...} \\
\search{Delicatessen celebrated American animator connection} \\
\info{Doc 1(Title: Delicatessen) items such as candy and mints are also usually available. Menus vary according to regional ethnic diversity. Although urban delis rely on ethnic meats (such as pastrami, corned beef and salami), supermarket delis rely on meats similar to their packaged meats (primarily ham, turkey and American bologna) ...
} \\
\hdashline
\think{I have found that the movie "Delicatessen" is associated with Jean-Pierre Jeunet, who is a celebrated film ... for the animator mentioned in the question.} \\
\search{Jacques Mathou Delicatessen American animator} \\
\info{Doc 1(Title: "Betty Blue") going to the publisher's house, slashes his face. Zorg induces him to drop charges by threatening him with violence, saying she is the only good thing in his life and she is all he has. Eddy's mother dies and the friends go to the funeral in Marvejols. There, Eddy asks Zorg and Betty if they will live in the dead woman's house and look after ...
} \\
\hdashline
\think{Upon further search, I have found that Jacques Mathou starred in "Delicatessen," but he is not an animator. The question seems to be focused on an American animator related to "Delicatessen." Since I lack this information, .."} \\
\answer{\textcolor{red}{Fail to answer}}\\
\hline
\textbf{w. E5}: \\
\think{I need to find out which celebrated American animator Delicatessen, the 1991 French movie, is in the style of. I'll search for it.} \\
\search{Delicatessen, the 1991 French movie in the style of which celebrated ..} \\
\info{Doc 1(Title: "Delicatessen (film)") Delicatessen (film) Delicatessen is a 1991 French post-apocalyptic black comedy film directed by Jean-Pierre Jeunet and Marc Caro, starring Dominique Pinon and Karin Viard. It was released in North America as ""presented by Terry Gilliam."" In a dilapidated apartment building in a post-apocalyptic country, food is in short supply ...
} \\
\hdashline
\think{I found out that Delicatessen is a 1991 French post-apocalyptic black comedy film directed by Jean-Pierre Jeunet and Marc Caro. Now I need to find out which celebrated American animator it is in the style of.} \\
\search{which celebrated American animator is Delicatessen in the style of} \\
\info{Doc 1(Title: "Joseph Barbera") Joseph Barbera Joseph Roland Barbera ( ; ; March 24, 1911 – December 18, 2006) was an American animator, director, producer, storyboard artist, and cartoon artist, whose film and television cartoon characters entertained millions of fans worldwide for much of the 20th century. He was born to Italian immigrants in New York City,} \\
\hdashline
\think{I found out that Delicatessen is in the style of Terry Gilliam.} \\
\answer{Terry Gilliam}\\
\hline
\end{tabular}
\end{table}

In Section \ref{sec:train-engine}, we have shown that LLMs trained with different search engines will learn different reasoning and search engine calling patterns.
In particular, the LLM trained with random noise will learn to not call the search engine since the return from the search engine is not informative; the LLM trained with a weak search engine (\textit{e.g.}, BM25) tend to call the search engine multiple times, while the LLM trained with a stronger search engine (\textit{e.g.}, E5) can call the search engine in a more reasonable pattern.

In this section, we would like to show case studies of LLMs trained with different search engine as below in Table \ref{apx:tab:case1} and Table \ref{apx:tab:case2}. 
The inference time search engine is the same to the training time search engine.

From these case studies, we find that during training, higher-quality search engines that provide more relevant information can encourage the agent to achieve its objectives with fewer search calls, as the retrieved content more effectively supports reasoning and decision-making. 
In contrast, lower-quality search engines that return less relevant information may lead the agent to either over-rely on its internal knowledge or issue multiple search queries to compensate for inadequate results.


\end{document}